\documentclass[runningheads]{llncs}

\usepackage{latexsym}
\usepackage{amssymb}
\usepackage{amsmath}
\usepackage{booktabs}
\usepackage{enumitem}
\usepackage{graphicx}
\usepackage{color}

\usepackage{times}
\usepackage{soul}
\usepackage{url}
\usepackage[hidelinks]{hyperref}
\usepackage[utf8]{inputenc}
\usepackage{graphicx}
\usepackage{amsmath}
\usepackage{booktabs}
\usepackage{algorithm}
\usepackage{algpseudocode}
\usepackage{amsmath}
\usepackage[switch]{lineno}
\usepackage{bm}
\usepackage{multirow}

\usepackage[utf8]{inputenc}
\usepackage{graphicx}
\usepackage{float}
\usepackage{algorithm}
\usepackage{algpseudocode}
\usepackage{tikz}
\usetikzlibrary{decorations.markings}
\usetikzlibrary{arrows}
\usetikzlibrary{shapes}
\usetikzlibrary{positioning}
\usepackage{pgfplots}
\pgfplotsset{compat=1.10}
\usetikzlibrary{shapes.geometric,arrows,fit,matrix,positioning}
\tikzset
{
    treenode/.style = {circle, draw=black, align=center, minimum size=1.1cm},
}
\newcommand{\size}[1]{|#1|}

\renewcommand{\vec}[1]{\bm{#1}}
\def\cnf{{\tt CNF}}
\def\dnf{{\tt DNF}}

\newcommand{\BibTeX}{B\kern-.05em{\sc i\kern-.025em b}\kern-.08em\TeX}

\begin{document}


\title{Leveraging Association Rules for Better Predictions\\ and Better Explanations}

\author{Gilles Audemard\inst{1}\orcidID{0000-0003-2604-9657} \and
Sylvie Coste-Marquis\inst{1}\orcidID{0000-0003-4742-4858} \and Pierre Marquis\inst{1,2}\orcidID{0000-0002-7979-6608} \and 
Mehdi Sabiri\inst{1}\orcidID{0009-0003-8642-9755} \and Nicolas Szczepanski\inst{1}\orcidID{0000-0001-7553-5657}}
\authorrunning{G. Audemard et al.}
%
\institute{Univ. Artois, CNRS, CRIL\\
\email{name@cril.fr}\\
\url{http://www.cril.fr} \and
Institut Universitaire de France 
}


\maketitle 
\begin{abstract}
We present a new approach to classification that combines data and knowledge. In this approach, data mining is used to derive association rules (possibly with negations) from data. Those rules are leveraged to increase the predictive performance of tree-based models (decision trees and random forests) used for a classification task.
They are also used to improve the corresponding explanation task through the generation of abductive explanations that are more general than those derivable without taking such rules into account.
Experiments show that for the two tree-based models under consideration, benefits can be offered by the approach in terms of predictive performance and in terms of explanation sizes.
\end{abstract}



\section{Introduction}

Hybrid AI is concerned with the design of more efficient AI systems based on both data and knowledge. 
Hybridizing data-driven techniques and knowledge-driven techniques can be achieved in many distinct ways and for a large number of purposes. 
Thus, beyond augmenting the predictive performance 
of the ML-based AI system one starts with, the use of symbolic information can be beneficial for better explaining the predictions that are made.
Such an interpretability issue (see e.g., \cite{Molnar19,DBLP:journals/inffus/ArrietaRSBTBGGM20}) is relevant to an XAI perspective that 
is crucial when dealing with critical applications \cite{DBLP:conf/iui/Gunning19}. 


In this paper, we present a new approach to classification where data and knowledge are combined.
Our objective is \emph{to determine to which extent the performance of a tree-based model (a decision tree or a random forest) used for a classification task can be enhanced as to inference and explanation by taking advantage of symbolic information under the form of association rules mined from the available data}. To achieve this goal, our approach exploits recent results concerning the correction of tree-based models \cite{DBLP:conf/ijcai/Coste-MarquisM21,coste-marquis23} and the efficient computation of abductive explanations for such models \cite{DBLP:conf/aaai/AudemardBBKLM22}.

Our approach basically consists of the following steps. Given a random forest $F$ (possibly reduced to a single decision tree) that has been learned from a dataset $D$, 
one first translates the instances from $D$ into instances over the Boolean conditions $X$ encountered in $F$, giving rise to a binarized dataset $D_b^F$. %
In general, the elements of $X$ do not represent independent conditions because they come from the same numerical or categorical attributes used to describe
the instances considered at start (those of $D$). As a consequence, a domain theory $Th$, represented as a Boolean formula on $X$, must be considered as well so as to make precise how the Boolean conditions in $X$ are logically connected (the pair $(F, Th)$ is a \emph{constrained decision-function} in the sense of \cite{DBLP:conf/aaai/GorjiR22}).
Then, 
a data mining algorithm is used to derive a (conjunctively-interpreted) set $A$ of association rules (possibly with negations) from $D_b^F$. Only rules with 100\% confidence and a non-null support are targeted. 
Among those rules, classification rules (alias class association rules - {\it CAR}, i.e., those rules from $A$ concluding about the membership to a class) forming a subset $A_c$ of $A$  are used to modify $F$ in such a way that the updated forest complies with the rules. 
Accordingly, primacy is given to the predictions that can be obtained using the classification rules over those coming from the random forest.
The rectification operation from  \cite{DBLP:conf/ijcai/Coste-MarquisM21,coste-marquis23} is used to update the forest. Algorithm \ref{algo} gives a pseudo-code of the operations that are achieved. 

\begin{algorithm}
  \caption{Computing a tree-based model rectified by a set of classification rules and a domain theory extended by association rules that are mined as well. }\label{algo} 
  \begin{algorithmic}
  \Require{a tabular dataset $D$.}\;
  \Ensure{a tree-based model $F^{A_c}$ complying with the classification rules from $A_c$ and an extended domain theory $Th_e$.}\;\\ 
  \State $(F, Th) \leftarrow learn(D)$\; 
  \State $D_b^F \leftarrow binarize(D, F)$\;
  \State $A \leftarrow mine(D_b^F)$\;
  \State $A_c \leftarrow select-CAR(A)$\;
  \State $F^{A_c} \leftarrow rectify(F, A_c)$\;
  \State $Th_e \leftarrow Th \wedge (A \setminus A_c))$\;\\
  \Return($(F^{A_c}, Th_e)$)\;
\end{algorithmic}
\end{algorithm}

One is also interested in 
deriving (local) abductive explanations for the predictions made. Thus, given an instance $\vec x$ and the corresponding classification $F(\vec x)$ to be explained,
one looks for a subset $t$ of the characteristics over $X$ used to represent $\vec x$ such that every instance $\vec x'$ that is covered by $t$ 
(i.e., such that $t$ is a subset of the characteristics of $\vec x'$) and that satisfies $Th_e = Th \wedge (A \setminus A_c)$
is classified by the rectified forest in the same way as $\vec x$. By considering instances that satisfies the \emph{extended domain theory} $Th_e$ (and not only 
$Th$), it is known that, in theory, more general explanations can be obtained  \cite{DBLP:conf/aaai/GorjiR22,DBLP:conf/aaai/YuISN023}. Since the explanations that are considered are based on Boolean 
conditions on $X$, minimum-size abductive explanations for an instance are among its most general explanations. Thus, the explanations computed using the
extended domain theory are in general shorter than those obtained when the initial domain theory is considered, and as a consequence, the extension of $Th$ achieved by considering in addition the rules from $A \setminus A_c$ leads to
abductive explanations that are typically easier to understand.


The contributions of the paper are as follows. 
We first show that it is possible to take advantage of the change operation for tree-based models, called \emph{rectification} \cite{DBLP:conf/ijcai/Coste-MarquisM21,coste-marquis23}, in order to incorporate the classification rules from $A_c$ into $F$. 
The rectification operation ensures that $F$, once rectified, classifies every instance on $X$ classified by a classification rule $R$ of $A_c$ in the way the rule $R$ asks for, while every other instance on $X$ is classified as required by $F$ before the rectification takes place. 
Especially, we show that the rectification of $F$ by $A_c$ an be achieved in an iterative way (i.e., on a rule-per-rule basis) 
provided that $A_c$ is non-conflicting (meaning that one cannot find in $A_c$ two classification rules with compatible premises but contradictory conclusions). In our approach, the conflict-freeness of $A_c$ is ensured by the process used for generating it. 

In order to derive valuable abductive explanations suited to tree-based models, a complexity challenge must be dealt with. Indeed, the generation of non-trivial abductive explanations for random forests is known to be computationally hard in general, even in the case when no domain theory is considered \cite{joao-ijcai21}. It is also known as hard 
for random forests with a single decision tree when a domain theory is taken into account \cite{DBLP:conf/ijcai/AudemardLMS24a}. To tackle this issue,
we show how to generalize the concept of \emph{majoritary reasons} introduced in \cite{DBLP:conf/aaai/AudemardBBKLM22} to decision trees and random forests based on Boolean conditions $X$ that are not independent but are connected through a domain theory on $X$; to ensure that the tractability of the computation of such abductive explanations, the inference relation used to reason from the domain theory is not full logical entailment but unit propagation.

A last contribution of the paper consists of an empirical evaluation of the approach in order to assess the benefits it offers. Interestingly, the experiments made show that the  objectives  of  improving  the  predictive performance of the classifiers and of diminishing the size of the explanations can be met. In a nutshell, for $12$ out of $13$ datasets used in the experiments, rectifying the random forest used by the classification rules that have been mined leads to slightly increase its predictive performance. The increase is small in general but it can exceed $10\%$.
As to the size of the abductive explanations that have been generated, the reduction achieved can be huge (more than 96\%) and effective for a large proportion of instances (up to 100\%), depending on the number of association rules that have been extracted. Depending on the dataset, the improvement observed is higher when the classifier used is a random forest or when it is a decision tree.


Additional empirical results and the code used in our experiments are available online at~\cite{swh-dir-588b380}.

\section{Formal Preliminaries}

\subsection{Decision tree, random forests, and classifiers} We consider a dataset $D$ consisting of classified instances from a binary classification problem and represented as tabular data. A finite set $\mathcal{A}$ 
of $p$ \emph{attributes} (aka \emph{features}) (where each attribute $A_i \in \mathcal{A}$ 
takes a value (called a \emph{characteristic}) in a domain $\mathit{Dom}_i$). 
Each attribute $A_i$ is numerical, categorical, or Boolean. An instance over $\mathcal{A}$ is a tuple from $\mathit{Dom}_1 \times \ldots \times \mathit{Dom}_p$.
The class of an instance in $D$ is made precise by the Boolean value of a specific column $y$ of $D$ (the instance is positive when $y$ takes value $1$ and negative when it takes value $0$).
Thus, a classified instance is a tuple from $\mathit{Dom}_1 \times \ldots \times \mathit{Dom}_p \times \{0, 1\}$.

When dealing with a binary classification problem, a \emph{decision tree} over $\mathcal{A}$ is a binary tree $T$, 
each of whose internal nodes (aka decision nodes) is labeled with a Boolean condition over $A_i \in \mathcal{A}$, 
and each leaf is labeled by a Boolean value denoting a class (positive or negative).
The value $T(\vec x)$ of $T$ on an input instance $\vec x$ is given by the label of the
leaf reached from the root, which is either a $1$-leaf (i.e., a leaf labelled by $1$) or a $0$-leaf (i.e., a leaf labelled by $0$).
The unique root-to-leaf path $p$ characterizing the leaf that is met is defined as follows:
at each decision node go to the left (resp. right) child if 
the Boolean condition labelling the node is evaluated to $0$ (resp. $1$) for $\vec x$. 
$\mathit{leaf}(p)$ denotes the leaf of $p$.

A \emph{random forest} over $\mathcal{A}$ is an ensemble $F = \{T_1,\cdots,T_m\}$, where each $T_i$ $(i \in [m])$ 
is a decision tree over $\mathcal{A}$. The value $F(\vec x)$ of $F$ on an input instance $\vec x$ is given by 
\begin{align*}
  F(\vec x) = 
    \begin{cases}
      1 & \mbox{ if } \frac{1}{m}\sum_{i=1}^m T_i(\vec x)  > \frac{1}{2} \\
      0 & \mbox{ otherwise.}
    \end{cases}
\end{align*}

The \emph{size} of $F$ is given by $\size{F} = \sum_{i=1}^m \size{T_i}$,
where $\size{T_i}$ is the number of nodes occurring in $T_i$. Clearly enough, any decision tree $T$ over $\mathcal{A}$ is equivalent to the random forest $F = \{T\}$ in the sense that $T(\vec x) = F(\vec x)$ for every instance $\vec x$ over $\mathcal{A}$. Therefore, in the rest of the paper, decision trees will also be viewed (wlog) as random forests with a single tree.

$X = \{x_1, \ldots, x_n\}$ denotes the set of Boolean conditions labelling decision nodes in $F$. 
A term (resp. a clause) on $X$ is a conjunction (resp. disjunction) of literals on $X$, i.e., of elements from $X$, possibly negated.
A \emph{positive literal} is an element $x \in X$, and a \emph{negative literal} is the negation $\overline{x}$ of an element $x \in X$.
When $\ell$ is a literal on $X$, its \emph{complementary literal}, noted $\sim \ell$, is given by $\overline{x}$ when $\ell = x$ is a positive literal,
and $x$ when $\ell = \overline{x}$ is a negative literal. A formula in \emph{disjunctive (resp. conjunctive) normal form} is a disjunction of terms
(resp. a conjunction of clauses).

Every root-to-leaf path $p$ of a decision tree $T_i$ is associated
with a term (noted $p$ as well to avoid heavy notations). For each decision node in $p$ labelled by $x \in X$, $x$ is a literal of $p$ when
the condition $x$ is evaluated to $1$ and $\overline{x}$ is a literal of $p$ otherwise.
It is well-known (see e.g.,  \cite{DBLP:conf/aaai/AudemardBBKLM22}) that
any decision tree $T_i$ on $X$ can be turned in linear time into an equivalent disjunction of consistent terms over $X$, noted $\dnf(T_i)$ and into 
an equivalent conjunction of non-valid clauses over $X$, noted 
$\cnf(T_i)$. Each term in $\dnf(T_i)$ corresponds to a $1$-path of $T_i$ and each clause in $\cnf(T_i)$ 
is the negation of a term describing a $0$-path of $T_i$.

Finally, $Th$ denotes a domain theory, i.e., a Boolean formula on $X$ that 
indicates how the Boolean conditions in $X$ are logically connected. Every instance $\vec x$ over $\mathcal{A}$ can be rewritten into a
(usually more general) instance over $X$ \cite{Audemardetal23} that is classified by $F$ in the same way as $\vec x$ so that
each $T_i$ in $F$ can be viewed as a decision tree over $X$, and $F$ can be viewed as a random forest over $X$. Each $T_i$ and $F$ itself
can thus be considered as Boolean formulas on $X$.
In the following, $\vec X$ denotes the set of all instances over $X$.
 
\begin{example}\label{ex:running}
As a matter of illustration, consider the following loan allocation problem. The goal is to determine whether a loan of $\$100k$ must be granted
to an applicant, described using three attributes from $\mathcal{A} = \{A, I, S\}$: his/her age $A$ (a numerical attribute), his/her annual income $I$ (in $\$k$, a numerical attribute), and his/her professional status $S$ (a categorical attribute). In the available dataset $D$, three values for $S$ are encountered (``unemployed (U)'',
``temporary position (TP)'', or ``permanent position''). The random forest $F = \{T_1, T_2, T_3\}$ given at Figure \ref{fig:RF} has been learned from 
$D$. $F$ is based on $7$ Boolean conditions: $X = \{x_1, \ldots, x_7\}$, where $x_1 = (A > 25)$, $x_2 = (A > 60)$, $x_3 = (I > 30)$, $x_4 = (I > 50)$, 
$x_5 = (S = U)$, $x_6 = (S = TP)$, and $x_7 = (S = PP)$. Those conditions are logically connected as given by the domain theory $Th = (x_2 \Rightarrow x_1) \wedge
(x_4 \Rightarrow x_3) \wedge (x_5 \Rightarrow \overline{x_6}) \wedge (x_5 \Rightarrow \overline{x_7}) \wedge (x_6 \Rightarrow \overline{x_7})$.
The implications $x_2 \Rightarrow x_1$ and $x_4 \Rightarrow x_3$ simply reflect that $60 > 25$ and $50 > 30$ (respectively). The remaining implications state that $S$ cannot take two distinct values in its domain at the same time. Here, the domain of the categorical attribute $S$ is considered open (i.e., we do not not assume that the only possible values for $S$ are $U$, $TP$, and $PP$).


The instance $(33, 52, PP)$ over $\mathcal{A}$ representing an applicant aged $32$, having $\$52k$ annual income, and a permanent position, 
corresponds to the instance $\vec x = (1, 0, 1, 1, 0,$ $0,$ $1)$ of $\vec X$.
$\vec x = (1, 0, 1, 1, 0, 0, 1)$ is more general than the instance $(33, 52, PP)$ one started with, in the sense that other
instances over $\mathcal{A}$ (e.g., $(48, 60, PP)$) also corresponds to $\vec x$. We can easily check that the instance $\vec x$ is such that
$F(\vec x) = 1$ (indeed, we have $T_1(\vec x) = 1$, $T_2(\vec x) = 0$, and $T_3(\vec x) = 1$).

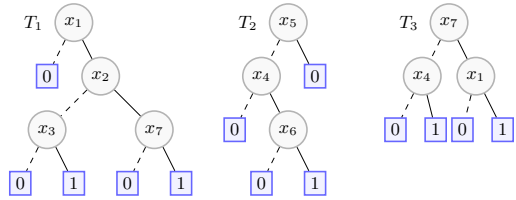
\begin{figure}[t]
    \centering
    \scalebox{0.75}{
          \begin{tikzpicture}[scale=0.475, roundnode/.style={circle, draw=gray!60, fill=gray!5, thick},
          squarednode/.style={rectangle, draw=blue!60, fill=blue!5, thick, minimum size=3mm}]
            \node[roundnode](root) at (2,7){$x_1$};
            \node at (0.5,7){$T_1$};
            \node[squarednode](n1) at (1,5){$0$};
            \node[roundnode](n2) at (3,5){$x_2$};
            \node[roundnode](n21) at (1,3){$x_3$};
            \node[roundnode](n22) at (5,3){$x_7$};  
            \node[squarednode](n211) at (0,1){$0$};
            \node[squarednode](n212) at (2,1){$1$};   
            \node[squarednode](n221) at (4,1){$0$};
            \node[squarednode](n222) at (6,1){$1$};   

            \draw[dashed] (root) -- (n1);
            \draw(root) -- (n2);
            \draw[dashed] (n2) -- (n21);
            \draw(n2) -- (n22);   
            \draw[dashed] (n21) -- (n211);
            \draw(n21) -- (n212);  
            \draw[dashed] (n22) -- (n221);
            \draw(n22) -- (n222);  

            \node[roundnode](root2) at (10,7){$x_5$};
            \node at (8.5,7){$T_2$};
            \node[roundnode](n2-1) at (9,5){$x_4$};  
            \node[squarednode](n2-2) at (11,5){$0$};
            \node[squarednode](n2-11) at (8,3){$0$};
            \node[roundnode](n2-12) at (10,3){$x_6$};  
            \node[squarednode](n2-121) at (9,1){$0$};
            \node[squarednode](n2-122) at (11,1){$1$};
            \draw[dashed] (root2) -- (n2-1);
            \draw(root2) -- (n2-2);
            \draw[dashed] (n2-1) -- (n2-11);
            \draw(n2-1) -- (n2-12);    
            \draw[dashed] (n2-12) -- (n2-121);
            \draw(n2-12) -- (n2-122);  
            
            \node[roundnode](root3) at (16,7){$x_7$};
            \node at (14.5,7){$T_3$};
            \node[roundnode](n3-1) at (15,5){$x_4$};  
            \node[roundnode](n3-2) at (17,5){$x_1$};  
            \node[squarednode](n3-11) at (14,3){$0$};
            \node[squarednode](n3-12) at (15.5,3){$1$};   
            \node[squarednode](n3-21) at (16.5,3){$0$};
            \node[squarednode](n3-22) at (18,3){$1$};     

            \draw[dashed] (root3) -- (n3-1);
            \draw(root3) -- (n3-2);
            \draw[dashed] (n3-1) -- (n3-11);
            \draw(n3-1) -- (n3-12);    
            \draw[dashed] (n3-2) -- (n3-21);
            \draw(n3-2) -- (n3-22); 
          \end{tikzpicture}
     }
        \caption{A random forest for a loan allocation problem. 
        The left (resp. right) child of any decision node labelled by $x_i$ corresponds to the assignment of $x_i$ to $0$ (resp. $1$).\label{fig:RF}}
        \end{figure}
\end{example}

%
Each instance  $\vec x$ of $\vec X$  can be considered as an interpretation on $X$ that satisfies $Th$. 
This interpretation can be represented by a (canonical) term $t_{\vec x}$ on $X$, formed by the set (interpreted as a conjunction) 
of the positive literals $x_i$ ($i \in [n]$), such that $\vec x_i = 1$ and by the negative literals $\overline{x_i}$ ($i \in [n]$) such that $\vec x_i = 0$.
$\top$ is the Boolean constant always true that evaluates to the Boolean value $1$ and $\bot$ is the Boolean constant always false that evaluates to the Boolean value $0$.
In the following, $\models$ denotes logical entailment and $\equiv$ logical equivalence.






\begin{definition}
A \emph{binary classifier} on $X$ is a mapping $C$ from $\vec X$ to the set of Boolean values $\{0, 1\}$.
$\vec x \in \vec X$ is a positive instance if $C(\vec x) = 1$  and a negative one if $C(\vec x) = 0$.
\end{definition}
Every ML model that is based on Boolean conditions (such as a decision tree or a random forest) and targets two classes only can be represented by a Boolean formula of the form $\Sigma_X \Leftrightarrow y$  where $\Sigma_X$  is a formula on $X$ and $y \not \in X$ is a propositional variable that denotes the class of positive instances. 
Indeed, it is sufficient to consider any $\Sigma_X$ such that $\vec x$ is a model of $\Sigma_X$ if and only if $\forall \vec x \in \vec X$, $C(\vec x) = 1$.

Accordingly, any binary classifier can be represented by a classification circuit in the sense of \cite{DBLP:conf/ijcai/Coste-MarquisM21}:
\begin{definition}
A \emph{classification circuit} $\Sigma$ on $X \cup \{y\}$ is a circuit equivalent to a formula of the form $\Sigma_X \Leftrightarrow y$ where $\Sigma_X$ is a Boolean formula on $X$.
\end{definition}
%
%
In the following, when $\Phi$ is a Boolean circuit or a formula on $X \cup \{y\}$ and $z$ is any variable from $X \cup \{y\}$, $\Phi(z)$ (resp. $\Phi(\overline{z})$) denotes the \emph{conditioning} of $\Phi$ by $z$ (resp. by $\overline{z}$).
$\Phi(z)$ (resp. $\Phi(\overline{z})$) is the circuit (or the formula) obtained by replacing in $\Phi$ any occurrence of $z$ by the Boolean constant $\top$ 
(resp. $\bot$).
When $\Sigma = \Sigma_X \Leftrightarrow y$ is a classification circuit on $X \cup \{y\}$, the set of models of $\Sigma(y)$ consists precisely of the models of $\Sigma_X$ and the set of models of $\Sigma(\overline{y})$ consists precisely of the counter-models of $\Sigma_X$. Finally, when $\vec x \in \vec X$ is an instance, $\Phi(\vec x)$ denotes the iterative conditioning of $\Phi$ by each literal of $t_{\vec x}$. Thus, $\vec x \in \vec X$ is classified positively (resp. negatively) by 
$\Sigma$ when $\Sigma(\vec x)$ is equivalent to $y$ (resp. $\overline{y}$).
%


\subsection{Implicants and abductive explanations} 
Given two formulas $\Phi$ and $Th$ on $X$, a term $t$ on $X$ is an \emph{implicant} of $\Phi$ modulo $Th$ iff $\Phi$ is a logical consequence of $t \wedge Th$. 
A term $t$ is an \emph{implicant} of $\Phi$ iff $\Phi$ is a logical consequence of $t$.  Thus, a term $t$ is an implicant of a formula $\Phi$ iff
$t$ is an implicant of $\Phi$ modulo a valid formula $\Phi$. 

Based on these notion of implicants, we can now make precise the notion of abductive explanation for an instance given a constrained decision-function based on a random forest \cite{DBLP:conf/aaai/GorjiR22}:

\begin{definition} 
Let $(F, Th)$ be a constrained decision-function where $F$ is a random forest on $X$ and $Th$ a Boolean formula on $X$.
Let $\vec x$ be an instance from $\vec X$ that satisfies $Th$ and is s.t. $F(\vec x) = 1$ (resp. $F(\vec x) = 0$).
\begin{itemize}
\item An {\em abductive explanation} for $\vec x$ 
given $(F, Th)$ is a (conjunctively-interpreted) set $t \subseteq t_{\vec x}$ such that $t$ is an implicant of $F$ modulo $Th$ (resp. $t$ is an implicant of $\overline{F}$ modulo $Th$).
\item A {\em subset-minimal abductive explanation} for $\vec x$ 
given  $(F, Th)$ is an abductive explanation $t$ for $\vec x$ given  $(F, Th)$ such that 
no proper subset of $t$ is an abductive explanation for $\vec x$ given  $(F, Th)$.
\end{itemize}
\end{definition} 


Abductive explanations $t$ provide subsets of characteristics of the instance $\vec x$ (thus, literals over $X$) that explain why $\vec x$ is classified by $F$ in the way it has been classified
(possibly, taking $Th$ into account).\footnote{As in \cite{IgnatievNM19} and unlike \cite{DBLP:journals/corr/abs-2012-11067}, in this paper, 
abductive explanations are not required to be minimal w.r.t. set inclusion. The notion of abductive explanations considered here corresponds to
so-called weak abductive explanations in \cite{DBLP:journals/corr/abs-2107-01654}.} 
Subset-minimal abductive explanations are also called sufficient reasons  \cite{DarwicheH20} and prime-implicant explanations \cite{ShihCD18}. 

\begin{example}[Example \ref{ex:running}, cont'ed]
A subset-minimal abductive explanation for $\vec x = (1, 0, 1, 1, 0, 0, 1)$ given  $(F, Th)$ is given by $t = x_1 \wedge \overline{x_2} \wedge x_4$.
This means that any applicant with age between 26 and 60 and annual income greater than $\$50k$ will be classified by $F$ in the same way
as $\vec x$ (i.e., as a positive instance), but if one of the three conditions (being older than 25, at least 60 years old, and having more than
$\$50k$ income) is relaxed, the same conclusion about the loan granted cannot be guaranteed.
\end{example}

\subsection{Association rules}In the following, we also need the notion of association rule:

\begin{definition}
An {\emph association rule} $R = b \Rightarrow h$ is a clause over $X \cup \{y\}$, where $y \not \in X$ denotes the class of positive instances. 
The body (aka premises) $b$ of such a rule always is a conjunction of literals over $X$ and the head (aka conclusion) $h$ of $R$ is a literal over $X \cup \{y\}$. When $h$ is a literal over $y$, the association rule $R = b \Rightarrow h$  is a {\emph classification rule} (aka a class association rule - CAR). 
\end{definition}

\begin{example}[Example \ref{ex:running}, cont'ed]
$(x_4 \wedge x_7) \Rightarrow y$ and $\overline{x_2} \Rightarrow \overline{x_7}$ are two association rules. 
$(x_4 \wedge x_7) \Rightarrow y$ is a classification rule.
\end{example}

The rules we consider may contain negations in their bodies and in their heads.  
Obviously, any (conjunctively-interpreted) set $A$ of association rules can also be viewed as a Boolean formula (in conjunctive normal form) on $X \cup \{y\}$.


Clearly enough, when it holds, a classification rule $R = b \Rightarrow y$ (resp. $R = b \Rightarrow \overline{y}$) indicates that each instance $\vec x \in \vec X$ satisfying $b$ has to be positively (resp. negatively) classified.
Indeed,  $R(\vec x) \equiv y$ (resp. $R(\vec x) \equiv \overline{y}$) is true.

Association rules can be generated using data mining techniques, including the seminal \textrm{Apriori} algorithm \cite{DBLP:conf/vldb/AgrawalS94}.
To do so, the classified instances over $\mathcal{A}$ from $D$ are first rewritten into classified instances over $X$ (the Boolean conditions used in $F$), giving rise to the binarized dataset $D_b^F$. In this dataset, each Boolean condition
$x \in X$ is used to give rise to two columns, one related to $x$, the other one to $\overline{x}$. This is useful to extract a set $A$ of association rules with negations
from $D_b^F$ using standard data mining algorithms.
Two key scores are typically used to assess the quality of an association rule. 
The \emph{support} of an association rule $R = b \Rightarrow h$ given a dataset $D_b^F$ is the number of instances (``transactions'') in $D_b^F$ satisfying $b \wedge h$ divided by the number of instances in $D_b^F$. The \emph{confidence} of a rule $R = b \Rightarrow h$ given a dataset $D_b^F$ is the number of instances in $D_b^F$ satisfying $b \wedge h$ divided by the number of instances satisfying $b$.

In general, the subset $A_c$ of $A$ consisting of classification rules derived from $D_b^F$ using a data mining algorithm corresponds to an \emph{incomplete} classifier since instances may exist that are not covered by the body of any rule from $A_c$. Furthermore, in the general case, $A_c$ can be \emph{conflicting} given the domain theory $Th$ on $X$. 
This means that one can find in $A_c$ two rules $R_1 = b_1 \Rightarrow h_1$ and $R_2 = b_2 \Rightarrow h_2$ such that $b_1 \wedge b_2 \wedge Th$ is consistent and $h_1\equiv \overline{h_2}$.
When $A_c$ contains conflicting rules
$R_1 = b_1 \Rightarrow h_1$ and $R_2 = b_2 \Rightarrow h_2$, one does not know how to classify instances $\vec x$ satisfying $b_1 \wedge b_2 \wedge Th$ since  the two rules give contradictory conclusions about the class of $\vec x$.



\section{Rectifying Decision Trees and Random Forests} 
\label{sec:rect}

To take advantage of the classification rules of $A_c$ (or, at least, those the user is sufficiently confident to them) to improve the predictive performance of $F$,
the \emph{rectification operation} \cite{DBLP:conf/ijcai/Coste-MarquisM21,coste-marquis23} can be used. To this purpose, class labels (reduced to $y$ or $\overline{y}$ here) are required to be explicit. Notably, rectification is not specific to tree-based models but it applies to \emph{classification circuits} $\Sigma$ on $X \cup \{y\}$. This is not an issue since when viewing a random forest $F$ as a Boolean formula over $X$, the formula $F \Leftrightarrow y$ is such a classification circuit $\Sigma$. 

Rectifying $F \Leftrightarrow y$ by $A_c$ then amounts to generate a classification circuit noted $(F \Leftrightarrow y) \star A_c$ such that for every instance $\vec x \in \vec X$,
if $A_c$ classifies $\vec x$, then $(F \Leftrightarrow y) \star A_c$ classifies $\vec x$ in the same way as $A_c$, else  $(F \Leftrightarrow y) \star A_c$ classifies $\vec x$ in the same way as $F$.
%
%
The rectified circuit $(F \Leftrightarrow y) \star A_c$ of $F \Leftrightarrow y$ by $A_c$ \cite{coste-marquis23} is characterized (up to logical equivalence) by
$(F \Leftrightarrow y) \star A_c \equiv F^{A_c} \Leftrightarrow y$ where 
$$F^{A_c} \equiv (F \wedge \neg (A_c(\overline{y}) \wedge \neg A_c(y))) \vee (A_c(y) \wedge \neg A_c(\overline{y})).$$

As explained in \cite{coste-marquis23}, the rationale of this characterization is as follows. For an instance $\vec x$ to be classified as positive by the rectified classification circuit,
it must be the case that either $A_c$ consistently asks for it (this corresponds to the disjunct $A_c(y) \wedge \neg A_c(\overline{y})$), or that the classification
circuit considered at start classifies $\vec x$ as positive, provided that $T$ does not consistently ask $\vec x$ to be classified as negative (this corresponds to the
disjunct $F \wedge \neg (A_c(\overline{y}) \wedge \neg A_c(y))$). When $A_c$ is conflict-free, every instance that can be classified using rules from $A_c$ is consistently classified by $A_c$.

In our approach, the rules in $A$ that are generated by the data mining algorithm $\mathit{mine}$ used in Algorithm \ref{algo} can be filtered by the user if needed (only those rules in which the user has sufficient trust must be used). A 100\% confidence score and a non-null support score is considered so that each rule
put forward by the data mining algorithm meets these conditions (the goal is generate only rules that are not contradicted by any piece of available evidence in $D_b^F$ since we want to use them as if they were true pieces of knowledge). 
The set $A$ of association rules is expected to be conflict-free but considering rules with a 100\% confidence score and a non-null support score is not enough to ensure it. 
In our approach, the lack of conflicts is ensured by the way the procedure $\mathit{mine}$ works:  rules are generated one by one, 
by decreasing support, and a generated rule $R$ is put into $A$ whenever it does not conflict with any of the rule $R'$ that precedes $R$ in the enumeration (provided that $R'$ has been kept so far and put into $A$).
The absence of conflicts makes it possible to rectify $F$ by $A_c$ in an iterative way, rule by rule
(the order with which the classification rules of $A_c$ are taken into account does not matter when $A_c$ is conflict-free).



Furthermore, in order to use XAI techniques developed so far for decision trees and random forests (especially, the computation of abductive explanations as presented in
Section \ref{sec:expl}), we would like the resulting classification circuit $(F \Leftrightarrow y) \star A_c$ to be represented as a decision tree when $F$ is a decision tree and as a random forest when $F$ is a random forest.
It is possible to ensure this property  without requiring additional heavy computational costs.
Indeed, in order to rectify a random forest $F$ by a classification rule $R$, it is enough to rectify every tree $T_i \in F$ by $R$.
Furthermore, rectifying a decision tree $T_i$ by a classification rule $R = b \Rightarrow h$ amounts to rectify every branch of $T_i$ by $R$, leading to a tree. Thus, all the branches of the trees in $F$ can be rectified by $R$ in parallel.

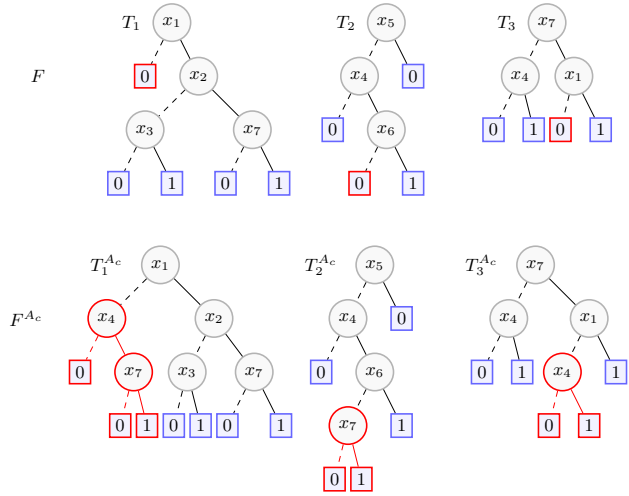
\begin{figure}[ht]
    \centering
        \scalebox{0.75}{
          \begin{tikzpicture}[scale=0.475, roundnode/.style={circle, draw=gray!60, fill=gray!5, thick},
          squarednode/.style={rectangle, draw=blue!60, fill=blue!5, thick, minimum size=3mm}]
          
            \node at (-3,5){$F$};
            
            \node[roundnode](root) at (2,7){$x_1$};
            \node at (0.5,7){$T_1$};
            \node[squarednode,draw=red](n1) at (1,5){$0$};
            \node[roundnode](n2) at (3,5){$x_2$};
            \node[roundnode](n21) at (1,3){$x_3$};
            \node[roundnode](n22) at (5,3){$x_7$};  
            \node[squarednode](n211) at (0,1){$0$};
            \node[squarednode](n212) at (2,1){$1$};   
            \node[squarednode](n221) at (4,1){$0$};
            \node[squarednode](n222) at (6,1){$1$};   

            \draw[dashed] (root) -- (n1);
            \draw(root) -- (n2);
            \draw[dashed] (n2) -- (n21);
            \draw(n2) -- (n22);   
            \draw[dashed] (n21) -- (n211);
            \draw(n21) -- (n212);  
            \draw[dashed] (n22) -- (n221);
            \draw(n22) -- (n222);  

            \node[roundnode](root2) at (10,7){$x_5$};
            \node at (8.5,7){$T_2$};
            \node[roundnode](n2-1) at (9,5){$x_4$};  
            \node[squarednode](n2-2) at (11,5){$0$};
            \node[squarednode](n2-11) at (8,3){$0$};
            \node[roundnode](n2-12) at (10,3){$x_6$};  
            \node[squarednode, draw=red](n2-121) at (9,1){$0$};
            \node[squarednode](n2-122) at (11,1){$1$};
            \draw[dashed] (root2) -- (n2-1);
            \draw(root2) -- (n2-2);
            \draw[dashed] (n2-1) -- (n2-11);
            \draw(n2-1) -- (n2-12);    
            \draw[dashed] (n2-12) -- (n2-121);
            \draw(n2-12) -- (n2-122);  
            
            \node[roundnode](root3) at (16,7){$x_7$};
            \node at (14.5,7){$T_3$};
            \node[roundnode](n3-1) at (15,5){$x_4$};  
            \node[roundnode](n3-2) at (17,5){$x_1$};  
            \node[squarednode](n3-11) at (14,3){$0$};
            \node[squarednode](n3-12) at (15.5,3){$1$};   
            \node[squarednode, draw=red](n3-21) at (16.5,3){$0$};
            \node[squarednode](n3-22) at (18,3){$1$};     

            \draw[dashed] (root3) -- (n3-1);
            \draw(root3) -- (n3-2);
            \draw[dashed] (n3-1) -- (n3-11);
            \draw(n3-1) -- (n3-12);    
            \draw[dashed] (n3-2) -- (n3-21);
            \draw(n3-2) -- (n3-22); 
            
            \node at (0,-1){};
            
          \end{tikzpicture}
          }

    \scalebox{0.75}{
          \begin{tikzpicture}[scale=0.475, roundnode/.style={circle, draw=gray!60, fill=gray!5, thick},
          squarednode/.style={rectangle, draw=blue!60, fill=blue!5, thick, minimum size=3mm}]
          
            \node at (-3,5){$F^{A_c}$};
          
            \node[roundnode](root) at (2,7){$x_1$};
            \node at (0,7){$T_1^{A_c}$};
            \node[roundnode, draw=red](n1) at (0,5){$x_4$};
            \node[squarednode, draw=red](n11) at (-1,3){$0$};       
            \node[roundnode, draw=red](n12) at (1,3){$x_7$};    
            \node[squarednode, draw=red](n121) at (0.5,1){$0$};  
            \node[squarednode, draw=red](n122) at (1.5,1){$1$};  
            \node[roundnode](n2) at (4,5){$x_2$};
            \node[roundnode](n21) at (3,3){$x_3$};
            \node[roundnode](n22) at (5.5,3){$x_7$};  
            \node[squarednode](n211) at (2.5,1){$0$};
            \node[squarednode](n212) at (3.5,1){$1$};   
            \node[squarednode](n221) at (4.5,1){$0$};
            \node[squarednode](n222) at (6.5,1){$1$};   

            \draw[dashed] (root) -- (n1);
            
            \draw[dashed, draw=red] (n1) -- (n11);
            \draw[draw=red] (n1) -- (n12);         
            \draw[dashed, draw=red] (n12) -- (n121);
            \draw[draw=red] (n12) -- (n122);    
            
            \draw(root) -- (n2);
            \draw[dashed] (n2) -- (n21);
            \draw(n2) -- (n22);   
            \draw[dashed] (n21) -- (n211);
            \draw(n21) -- (n212);  
            \draw[dashed] (n22) -- (n221);
            \draw(n22) -- (n222);  

            \node[roundnode](root2) at (10,7){$x_5$};
            \node at (8,7){$T_2^{A_c}$};
            \node[roundnode](n2-1) at (9,5){$x_4$};  
            \node[squarednode](n2-2) at (11,5){$0$};
            \node[squarednode](n2-11) at (8,3){$0$};
            \node[roundnode](n2-12) at (10,3){$x_6$};  
            \node[roundnode,draw=red](n2-121) at (9,1){$x_7$};
            \node[squarednode,draw=red](n2-1211) at (8.5,-1){$0$};
            \node[squarednode,draw=red](n2-1212) at (9.5,-1){$1$};
                        
            \node[squarednode](n2-122) at (11,1){$1$};
            \draw[dashed] (root2) -- (n2-1);
            \draw(root2) -- (n2-2);
            \draw[dashed] (n2-1) -- (n2-11);
            \draw(n2-1) -- (n2-12);    
            \draw[dashed] (n2-12) -- (n2-121);
            \draw(n2-12) -- (n2-122);  
            \draw[dashed,draw=red] (n2-121) -- (n2-1211);
            \draw[draw=red](n2-121) -- (n2-1212);             
            
            \node[roundnode](root3) at (16,7){$x_7$};
            \node at (14,7){$T_3^{A_c}$};
            \node[roundnode](n3-1) at (15,5){$x_4$};  
            \node[roundnode](n3-2) at (18,5){$x_1$};  
            \node[squarednode](n3-11) at (14,3){$0$};
            \node[squarednode](n3-12) at (15.5,3){$1$};   
            \node[roundnode,draw=red](n3-21) at (17,3){$x_4$};
            \node[squarednode,draw=red](n3-211) at (16.5,1){$0$};  
            \node[squarednode,draw=red](n3-212) at (18,1){$1$};              
            \node[squarednode](n3-22) at (19,3){$1$};     

            \draw[dashed] (root3) -- (n3-1);
            \draw(root3) -- (n3-2);
            \draw[dashed] (n3-1) -- (n3-11);
            \draw(n3-1) -- (n3-12);    
            \draw[dashed] (n3-2) -- (n3-21);
            \draw(n3-2) -- (n3-22); 
            \draw[dashed,draw=red] (n3-21) -- (n3-211);
            \draw[draw=red](n3-21) -- (n3-212); 
          \end{tikzpicture}
     }
        \caption{The random forest $F^{A_c}$ obtained by rectifying $F$ by $A_c = \{ (x_4$ $\wedge$ $x_7) \Rightarrow y\}$.
        The modifications achieved w.r.t. $F$ are printed in \textcolor{red}{red}. \label{fig:RF-rectified}}
        \end{figure}

More precisely, rectifying $T_i$ by $R = b \Rightarrow h$ amounts to update only those root-to-leaf paths $p$ of $T_i$ 
such that $p \wedge b \wedge Th$ is consistent and the leaf of $p$ conflicts with the conclusion $h$ of the rule (i.e., when $h = y$
and the leaf of $p$ is a $0$-leaf and when $h = \overline{y}$ and the leaf of $p$ is a $1$-leaf). 
Let $\mathit{patch}(p, R)$ the {\it comb-shaped tree} with its main branch labelled by conditions from $b \setminus p$ leading to a $0$-leaf if
$h =  \overline{y}$ and to a $1$-leaf if $h =  y$, while every other branch of the comb-shaped tree is labelled by $\mathit{leaf}(p)$.
A rectification of $p$ by $R$ can be obtained by replacing the leaf node of $p$ by $\mathit{patch}(p, R)$.

\begin{example}[Example \ref{ex:running}, cont'ed]
Suppose that the data mining algorithm has produced a unique classification rule with confidence 100\% and a non-null support, namely 
$R = (x_4 \wedge x_7) \Rightarrow y$. This rule is not a logical consequence of the classification circuit $F \Leftrightarrow y$.
Indeed, the instance $\vec x = (0, 0, 1, 1, 0, 0, 1)$ is such that $F(\vec x) = 0$ while $\vec x$ is classified as a positive instance by $R$.
If $R$ is considered reliable enough by the user, $F \Leftrightarrow y$ can be rectified by $A_c = \{R\}$.
The resulting random forest $F^{A_c}$ is given at Figure \ref{fig:RF-rectified}. In detail, with $b = x_4 \wedge x_7$ and $h = y$:
\begin{itemize}
    \item The unique branch of $T_1$ that needs to be rectified is the one corresponding to $\overline{x_1}$ since it ends with a $0$-leaf while $h = y$. The two other branches of $T_1$ ending with a $0$-leaf are associated with terms $p$ (namely, $x_1 \wedge \overline{x_2} \wedge \overline{x_3}$ and $x_1 \wedge \overline{x_2} \wedge \overline{x_7}$) such that $p \wedge b \wedge Th$ is inconsistent (keep in mind that 
    $x_4 \Rightarrow x_3$ is part of $Th$). Thus, the $0$-leaf node corresponding to the path $\overline{x_1}$ in $T_1$ is replaced by the comb-shaped tree $\mathit{patch}(\overline{x_1}, R)$ in $T_1^{A_c}$. This comb-shaped tree simply ensures that a $1$-leaf is reached when $x_4 \wedge x_7$ holds, while a $0$-leaf (i.e., $\mathit{leaf}(\overline{x_1})$) is reached otherwise.
    \item The unique branch of $T_2$ that needs to be rectified is the one corresponding to $\overline{x_5} \wedge x_4 \wedge \overline{x_6}$ since it ends with a $0$-leaf while $h = y$. The two other branches of $T_2$ ending with a $0$-leaf are associated with terms $p$ (namely, $\overline{x_5} \wedge \overline{x_4}$ and $x_5$) such that $p \wedge b \wedge Th$ is inconsistent (keep in mind that 
    $x_5 \Rightarrow \overline{x_7}$ is part of $Th$). Thus, the $0$-leaf node corresponding to the path $\overline{x_5} \wedge x_4 \wedge \overline{x_6}$ in $T_2$ is replaced by the comb-shaped tree $\mathit{patch}(\overline{x_5} \wedge x_4 \wedge \overline{x_6}, R)$ in $T_2^{A_c}$. This comb-shaped tree simply ensures that a $1$-leaf is reached when $x_4 \wedge x_7$ holds, while a $0$-leaf (i.e., $\mathit{leaf}(\overline{x_5} \wedge x_4 \wedge \overline{x_6})$) is reached otherwise. In the comb-shaped tree $\mathit{patch}(\overline{x_5} \wedge x_4 \wedge \overline{x_6}, R)$, there is no need to repeat the condition $x_4$ since it belongs to the path $\overline{x_5} \wedge x_4 \wedge \overline{x_6}$.
    \item Finally, the unique branch of $T_3$ that needs to be rectified is the one corresponding to $x_7 \wedge \overline{x_1}$ since it ends with a $0$-leaf while $h = y$. The other  branch of $T_3$ ending with a $0$-leaf is associated with the term $p = \overline{x_7} \wedge \overline{x_4}$ such that $p \wedge b \wedge Th$ is inconsistent. Thus, the $0$-leaf node corresponding to the path $x_7 \wedge \overline{x_1}$ in $T_3$ is replaced by the comb-shaped tree $\mathit{patch}(x_7 \wedge \overline{x_1}, R)$ in $T_3^{A_c}$. This comb-shaped tree simply ensures that a $1$-leaf is reached when $x_4 \wedge x_7$ holds, while a $0$-leaf (i.e., $\mathit{leaf}(x_7 \wedge \overline{x_1})$) is reached otherwise. In the comb-shaped tree $\mathit{patch}(x_7 \wedge \overline{x_1}, R)$, there is no need to repeat the condition $x_7$ since it belongs to the path $x_7 \wedge \overline{x_1}$.
\end{itemize}
\end{example}

The rectification of $F$ by $R$ can be achieved in time $\mathcal{O}(|F| \cdot |R|)$, leading to a model $F^{A_c}$ of size upper bounded by $\mathcal{O}(|F| \cdot |R|)$. Notably, 
the branches of each resulting tree $T_i^{A_c}$ in $F^{A_c}$ can be {\it simplified} using $Th$: from bottom to top, starting from the leaf of a branch $p$ up to the root of the tree $T_i$,
an arc of $p$ can be removed when the literal $\ell$ labelling it is a logical consequence of $(p \setminus \{\ell\}) \wedge Th$. In addition, any internal node of $T_i^{A_c}$ having
a left subtree identical to its right subtree can be replaced by one of its two subtrees. Though such a simplification process would let unchanged the rectified model $F^{A_c}$ considered in the running example, in the general case it may have a significant effect on the size of the resulting model, leading sometimes to a model that is smaller than the one $F$ we started with. To illustrate this point, consider the running example again: it is easy to verify that rectifying $T_3$ by $(\overline{x_7} \wedge \overline{x_4}) \Rightarrow y$ and simplifying the resulting tree would lead to a tree smaller than $T_3$ itself (we would get a tree with $5$ nodes instead of $7$ nodes for $T_3$).

\section{Deriving Abductive Explanations} 
\label{sec:expl}

Let us turn now to the second problem tackled, i.e., computing better explanations by leveraging association rules.
A first important remark is that, though no consensus exists  about what a ``good'' explanation should be \cite{Miller19,Molnar19}, 
many criteria for evaluating explanations (and/or the XAI methods used to produce them)
have been pointed out \cite{Nautaetal23}. Some of those criteria are antagonistic, so trade-offs must be considered.
Among other criteria, \emph{correctness} indicates to which extent explanations capture the actual behaviour of the AI system (and not the one of a surrogate model). 
\emph{Compactness} concerns the size of the explanations (shorter explanations are usually easier to interpret). 
When dealing with (local) abductive explanations based on Boolean conditions (as it is the case in this paper), 
the size of the explanations is also related to their \emph{generality}, i.e., the set of instances covered by the explanation. Indeed, in this case, 
minimum-size abductive explanations are among the subset-minimal abductive explanations, i.e., those that are
as general as possible. \emph{Coherence} is about whether explanations comply with the domain knowledge, while \emph{controllability}
refers to the possibility for a user to influence the explanations that are provided.

Unlike other notions of explanations based on feature attribution techniques (especially Shapley values \cite{DBLP:journals/cacm/MarquesSilvaH24,DBLP:journals/ijar/HuangM24}), 
abductive explanations $t$ for an instance $\vec x$ given $(C, Th)$, are formal explanations that are {\emph correct by design}: it is ensured that for any feasible 
instance $\vec x'$ (i.e., an instance that satisfies $Th$), if $\vec x'$ is covered by $t$, then $C(\vec x') = C(\vec x)$.
Stated otherwise, $t$ being true is enough to explain the way $\vec x$ has been classified by $C$. The coherence criterion is also
met given that pieces of domain knowledge (given as a domain theory) are exploited in the computation of explanations.

Computing an abductive explanation for $\vec x \in \vec X$ given $(C, Th)$ is an easy task 
when no requirement are imposed about the generality of the explanation that is generated. Indeed, $t_{\vec x}$ is a (trivial) abductive 
explanation for $\vec x \in \vec X$ given $(C, Th)$. The direct reason for $\vec x$, i.e., the set of characteristics of $t_{\vec x}$ that can be found in the unique path of $C$ compatible with $t_{\vec x}$ when $C$ is a decision tree, and the union of all those characteristics over the trees of $C$ when $C$ is a random forest, is an alternative abductive 
explanation for $\vec x \in \vec X$ given $(C, Th)$. However, such a direct reason (aka the path-restricted explanation for $\vec x$ when $C$ is a decision tree) may contain many redundant conditions that are not present in subset-minimal abductive explanations \cite{DBLP:journals/jair/IzzaIM22}.

The difficulty is to thus to find more general explanations, especially 
subset-minimal abductive explanations or even minimum-size abductive explanations.
Going a step further in this direction requires to address \emph{a complexity issue} because generating subset-minimal abductive explanations
or minimum-size abductive explanations is computationally difficult.
The presence of a domain theory makes (in general) the problem harder. Thus, while the generation of a subset-minimal abductive explanation for an instance given a decision tree can be achieved in polynomial time when no domain theory is considered, the problem becomes (in general) {\sf NP}-hard when a domain theory must be taken into account \cite{DBLP:conf/ijcai/AudemardLMS24a}. 
On the other hand, the problem of generating a subset-minimal abductive explanation for $\vec x$ given a random forest is intractable (${\sf DP}$-hard) \cite{joao-ijcai21},  even when no domain theory is taken into account.
Of course, generating  minimum-size abductive explanations is at least as hard as generating subset-minimal abductive explanations since every minimum-size abductive explanation necessarily is a subset-minimal abductive explanation. 

To deal with such a complexity issue, 
the concept of \emph{majoritary reason} \cite{DBLP:conf/aaai/AudemardBBKLM22} has been pointed out as a valuable trade-off 
in terms of tractability of the computation and generality. Majoritary reasons are abductive explanations (thus satisfying the correctness criterion above)
that can be computed in polynomial time in the case of random forests and that coincides with subset-minimal abductive explanations in the case of decision trees (i.e., random forests with a single tree). Furthermore, even if majoritary reasons may contain irrelevant characteristics in general,
in practice one can often derive majoritary reasons
that are shorter than subset-minimal abductive explanations  \cite{DBLP:conf/aaai/AudemardBBKLM22}.
Formally, majoritary reasons have been defined as follows \cite{DBLP:conf/aaai/AudemardBBKLM22}:

\begin{definition}
Let $F = \{T_1, \ldots, T_m\}$ be a random forest over $X$ and $\vec x \in \vec X$. 
A \emph{majoritary reason} for $\vec x$ given $F$ is a term $t$ covering $\vec x$ (i.e., $t$ a subset of $t_{\vec x}$), such that $t$ 
is an implicant of at least $\lfloor \frac{m}{2} \rfloor +1$ decision trees $T_i$ (resp. $\neg T_i$) if $F(\vec x) = 1$ (resp. $F(\vec x) = 0$), 
and for every $\ell \in t$, $t \setminus \{\ell\}$ does not satisfy this last condition.
 \end{definition}
 
In the following, we show how the concept of majoritary reason for random forests $F$ presented in \cite{DBLP:conf/aaai/AudemardBBKLM22}
can be extended to the case when a domain theory $Th_e$ is considered. 
The extended domain theory used here includes constraints that encode the logical connections between Boolean conditions used in $F$,
possibly completed by association rules that have been derived from $D_b^F$.

\begin{definition}
Let $F = \{T_1, \ldots, T_m\}$ be a random forest over $X$, $Th_e$ a domain theory on $X$, and $\vec x \in \vec X$.
A \emph{majoritary reason} for $\vec x$ given $(F, Th_e)$ is a term $t$ covering $\vec x$ (i.e., $t$ a subset of $t_{\vec x}$), such that $t$ 
is an implicant modulo $Th_e$ of at least $\lfloor \frac{m}{2} \rfloor +1$ decision trees $T_i$ (resp. $\neg T_i$) if $F(\vec x) = 1$ (resp. $F(\vec x) = 0$), 
and for every $\ell \in t$, $t \setminus \{\ell\}$ does not satisfy this last condition.
 \end{definition}
 
The problem with this definition is that it leads to a concept of reason that cannot be computed in polynomial time (unless {\sf P = NP}). Indeed, in the general case, no constraint bears
on the association rules that can be generated, so they can be any clauses. As a consequence, testing whether $t$ is an implicant modulo $Th$ of a tree $T_i$
is a {\sf coNP}-complete problem, which precludes the existence of polynomial-time algorithms for generating majority reasons.

%


To deal with this problem while taking into account within the extended domain theory $Th_e$ the set of all the association rules produced by the data mining algorithm used, 
the approach we followed consists in \emph{weakening the inference relation used to reason}.
Let us remind that \emph{unit resolution} is an inference rule allowing to derive a clause $\delta$ on $X$ from a literal $\ell$ on $X$ (called a unit clause) and
a clause $\sim \ell \vee \delta$ on $X$. A literal $\ell$ is \emph{derivable by unit propagation} from a domain theory $Th_e$ in 
conjunctive normal formal (CNF), noted $Th_e \vdash_1 \ell$ iff there exists a finite sequence of clauses $\delta_1, \ldots, \delta_k$ on $X$ such that $\delta_k = \ell$ and every clause
$\delta_i$ ($i \in [k]$) in the sequence is a clause of $Th_e$ or can be obtained by applying the unit resolution rule to two clauses $\delta_a, \delta_b$ of the sequence
such that $a < i$ and $b < i$. The set of all the literals on $X$ that are derivable by unit propagation (UP) from $Th_e$ can be computed in time linear
in the size of $Th_e$ (see e.g., \cite{Dalal1992a,DBLP:journals/jar/ZhangS00}). Every literal from this set is a logical consequence of $Th$ (in symbols, if $Th \vdash_1 \ell$ then $Th \models \ell$), but the converse does not hold in general.

Let  $\Phi$ be a formula on $X$ in CNF that does not contain any valid clause. When $t$ is a consistent term, 
$t$ is an implicant of $\Phi$ iff every clause $\delta$ of $\Phi$ contains a literal $\ell$ of $t$.
Given a domain theory $Th$ in conjunctive normal term, let us say that a term $t$ on $X$ is a \emph{UP-implicant} of $\Phi$ iff for every clause $\delta$ of $\Phi$,
$\delta$ contains a literal $\ell$ that belongs the set of literals derivable by unit propagation from $t \wedge Th$. 
Clearly enough, if $t$ is a UP-implicant of $\Phi$ given $Th$, then $t$ is an implicant of $\Phi$ modulo $Th$, but the converse does not hold in general.
For instance, $\overline{b}$ is a logical consequence of $t \wedge Th$, with $t = \top$ (the empty term) and 
$Th = (a \vee \overline{b}) \wedge (\overline{a} \vee \overline{b})$ but $\overline{b}$ is not derivable by unit propagation from $t \wedge Th$.
Based on this notion of UP-implicant, a corresponding notion of majoritary reason can be defined as well:

\begin{definition}
Let $F = \{T_1, \ldots, T_m\}$ be a random forest over $X$, $Th_e$ a domain theory on $X$, and $\vec x \in \vec X$. 
A \emph{UP-majoritary reason} for $\vec x$ given $(F, Th_e)$ is a term $t$ covering $\vec x$, such that $t$ 
is a UP-implicant given $Th_e$ of at least $\lfloor \frac{m}{2} \rfloor +1$ decision trees $T_i$ (resp. $\neg T_i$) if $F(\vec x) = 1$ (resp. $F(\vec x) = 0$), 
and for every $\ell \in t$, $t \setminus \{\ell\}$ does not satisfy this last condition.
 \end{definition}


The tractability of generating UP-majoritary reasons lies in the fact that they can be computed using a simple greedy algorithm equipped with
unit propagation instead of full logical entailment. 

\begin{proposition}\label{prop:greedy}
Given a  random forest $F$ over a set of Boolean conditions $X$, a domain theory $Th$ over $X$ in conjunctive normal form, 
and an instance $\vec x \in \vec X$, a UP-majoritary reason for $\vec x$ given $(F, Th)$ can be computed in
time polynomial in the size of the input.
\end{proposition}


Considering distinct orderings over the literals of $t = t_{\vec x}$ within the greedy algorithm may easily lead it to generate distinct majoritary reasons.
Since the greedy algorithm is efficient, several orderings can be tested and finally, for taking the compactness criterion into account, 
one of the shortest majoritary reasons among those that have been produced can be retained. Furthermore, the fact that the greedy  algorithm is order-driven
can be exploited to focus the search for majoritary reasons to explanations that preferentially contain (or not contain) some characteristics
(to do so, it is enough to order the literals of $t_{\vec x}$ from the least preferred to the most preferred). The ordering used thus reflects some user preferences
(this is a first way to meet the controllability criterion).

In our approach, association rules that are not classification rules  
are added to the initial domain theory $Th$ composed of clauses that encode the logical connections between Boolean conditions used in $F$. 
This leads to an extended domain theory $Th_e$ in conjunctive normal form.
As it was the case for the classification rules considered in Section \ref{sec:rect}, 
only rules having a 100\% confidence score and a non-null support are extracted by the data mining algorithm (again, the goal is generate only rules that are not contradicted 
by any piece of available evidence in $D_b^F$ since we want to use them as if they were true pieces of knowledge). 
Among them, only those rules in which the user is trustful enough can be kept (this is another way to meet the controllability criterion).
%

Interestingly, extending $Th$ to $Th_e$ may lead to the generation of more general explanations. Indeed, whenever a domain $Th_e$ is at least as strong as another domain theory $Th$ from a logical point of view (here, because $Th$ has been completed by association rules to get $Th_e$), the formula $Th_e \Rightarrow F$ is a logical consequence of the formula $Th \Rightarrow F$ (and similarly for $\overline{F}$ instead of $F$). As a consequence, for every implicant $t$ of $F$ modulo $Th$, there exists an implicant $t'$ of $F$ modulo $Th_e$ such that $t'$ is implied by $t$. Similarly, for every UP-implicant $t$ of $F$ given $Th$, 
since the set of literals derivable by unit propagation from $Th$ is a subset of the set of literals derivable by unit propagation from $Th_e$ when $Th \subseteq Th_e$,
there exists an implicant $t'$ of $F$ given $Th_e$ such that $t'$ is a logical consequence of $t$.  In both cases, $t'$ is at least as general as $t$. 

\begin{example}[Example \ref{ex:running}, cont'ed]
Consider the rectified random forest $F^{A_c}$  at Figure \ref{fig:RF-rectified} and 
$\vec x = (1, 0, 1, 1, 0, 0, 1)$.
We have $F^{A_c}(\vec x) = 1$. 
We can check that $t = x_1 \wedge \overline{x_2} \wedge x_4$ is a UP-majoritary reason for $\vec x$ given $(F^{A_c}, Th)$, and even 
a subset-minimal abductive explanation for $\vec x$ given $(F^{A_c}, Th)$.
Suppose that a unique association rule $R = (x_1 \wedge \overline{x_2}) \Rightarrow x_4$ with a 100\% confidence score and a non-null support 
has been extracted 
and that the user is confident that this rule is correct. 
Then $R$ can be added to $Th$, to give the extended domain theory $Th_e$. 
$t$ is not a UP-majoritary reason for $\vec x$ given $(F^{A_c}, Th_e)$ but the more general term $t' = x_1 \wedge \overline{x_2}$ is such a reason.
\end{example}


\section{Experiments}\label{sec:expe}

\subsection{Empirical protocol} 
Table \ref{tab:description} shows the description of the $13$ datasets $D$ used in our experiments and indicates the repositories where they can be found: 
UCR,\footnote{\url{www.timeseriesclassification.com}}
UCI,\footnote{\url{https://archive.ics.uci.edu/ml/index.php}} and openML.\footnote{\url{https://www.openml.org/}}

Categorical features have been treated as nominal variables and encoded as arbitrary numbers. As to numerical features, no explicit preprocessing was performed: these features were binarized on-the-fly by the decision tree and/or random forest algorithms used for learning (we used the latest version of the \texttt{Scikit-Learn} library~\cite{scikit-learn}).  
All hyperparameters of the learning algorithms were set to their default values (no preset depth for the trees and $100$ trees in the random forests).

For every dataset $D$, a repeated random sub-sampling cross validation process
has been achieved. $D$ has been split $10$ times into two subsets: a training set gathering 70\% of the available instances from $D$ (chosen at random) and a test set composed of the remaining 30\% of instances from $D$. For each partition, a decision tree (resp. a random forest) $F$ has been learned from the corresponding training set of the partition and the performance of the classifier has been evaluated on the corresponding test set. This performance scores (as well as the sizes of the classifiers and other characteristics of them) are averaged over the $10$ models $F$ that have been learned.

\begin{table}[t]
\caption{\label{tab:description}Description of the empirical setting.}\vspace{0.4cm}
\centering

\begin{tabular}{ l r r r  r  r r r r}
\toprule
Dataset&$|D|$ & $|\mathcal{A}|$ & $X_{RF}$ &$X_{DT}$ & $R_{RF}$ & $R_{DT}$&Repository\\
\midrule
arrowhead\_0\_vs\_1 & 146 & 249 & 611.5 &6.8 &100.0&16.6 &UCR \\
arrowhead\_0\_vs\_2 & 146 & 249 & 341.4 &5.4 &100.0&6.8 &UCR \\
arrowhead\_1\_vs\_2 & 130 & 249 & 593.0 &6.7 &100.0 &10.8 &UCR\\
australian & 690 & 38 & 1299.8 &47.5 &100.0&99.2 &openML \\
balance & 576 & 4 & 28.0 &19.4 &30.3&18.8 &UCI \\
biodegradation & 1055 & 41 & 4320.4 &76.6 &100.0&100.0 &openML \\
breast-tumor & 286 & 37 & 112.1 &38.8 &100.0&100.0 &openML \\
cleveland & 303 & 22 & 551.0 &27.3 &100.0&40.7 &openML \\
cnae & 1080 & 856 & 75.8 &45.8 &99.6&14.5 &UCI \\
compas & 6172 & 11 & 68.0 &49.1 &97.9&70.6 &openML \\
contraceptive & 1473 & 21 & 108.6 &75.3 &100.0& 100.0&UCI \\
divorce & 170 & 54 & 83.3 &1.7 &100.0&2.8 &UCI \\
german & 1000 & 58 & 406.5 &34.4 &100.0 &100 &UCI\\
\bottomrule
\end{tabular}
\end{table}

Then, for each dataset $D$ and classifier $F$, an associated binarized dataset $D_b^F$ was generated using the Boolean conditions occurring in $F$. A set $A$ of association rules $R$ with negations was derived from $D_b^F$ using the ad-hoc data mining algorithm $\mathit{mine}$. 
Only rules $R$  having a confidence of 100\%, a non-null support and that are not falsified by any element $(\vec x, \ell)$ of 
$D_b^F$ have been extracted from the training set of $D_b^F$. This means that whenever the body $b$ of $R$ is satisfied by $\vec x$, the head $h$ of $R$ agrees with the label $\ell$ of $\vec x$. Those conditions are considered to limit the risk that the association rules that are mined do not hold.
No specific properties (e.g., closed itemset) were targeted for the antecedent of the classification rules. Finally, the rules $R$ of $A$ were generated by decreasing support\footnote{Note that generating the rules by decreasing lift would not be discriminant enough. Indeed, the lift of a rule can be computed by dividing its confidence by the unconditional probability of its head, so when dealing with rules having a confidence of 100\%, all the rules with the same head have the same lift.}, and, once generated, a rule was kept if and only if it does not conflict with a rule that has been generated and kept before.
A timeout of $3600$ seconds was considered for the generation of classification rules of $A_c$, with a limit of $100$ rules to maintain the size of the trees in $F^{A_c}$ ``small enough''. A timeout of $3600$ seconds was considered as well for the generation of other association rules, i.e., those from $A \setminus A_c$, with various limits in the number of rules (100 to 100 000) in order to evaluate the impact of the logical strength of $Th_e$ on the size of the explanations.
Only association rules of size $2$ and $3$ have been mined by $\mathit{mine}$ 
(indeed, most general rules are the most interesting ones).

Since some of the datasets used in the experiments are unbalanced, the classification performance of each model on a dataset was measured using the average F-score, average G-mean, and average AUC score of the corresponding classifiers over the $10$ test sets
before any rectification, and once the classifiers have been rectified by the corresponding set $A_c$ of classification rules. 

In order to measure the impact of using the association rules from $A \setminus A_c$  on the size of explanations, $100$ instances have been selected from each test set and those violating any association rule from $A \setminus A_c$ have been discarded. Then for every remaining instance $\vec x$, UP-majoritary reasons for $\vec{x}$ given $(F, Th)$ were generated using a greedy algorithm that starts with $t_{\vec x}$ and exploits an (elimination) ordering over the features of $t_{\vec x}$. $100$ orderings per $\vec x$ have been selected at random, leading possibly to $100$ distinct reasons for $\vec x$. A shortest reason among those generated has been retained, and the mean size of those shortest reasons 
over the set of instances has been computed. The mean number of instances (out of $100$) for which a size decrease has been observed has also been computed. Finally, the same task has been achieved but considering this time
the extended domain theory $Th_e = Th \wedge (A \setminus A_c)$ instead of $Th$.

%

In addition to the specification of the repositories where the datasets used come from, Table \ref{tab:description} provides different statistics. Column $|D|$ represents the number of instances in  dataset $D$. $|\mathcal{A}|$ represents the number of primitive features used to describe the instances of $D$, $X_{RF}$ (resp. $X_{DT}$) represents the average number of features in 
the $10$ binarized datasets $D_b^F$, obtained from $D$ by using the Boolean conditions in the $10$ random forests (resp. in the $10$ decision trees) that have been generated. $R_{RF}$ and $R_{DT}$ represents the average number of classification rules extracted from the $10$ binarized datasets 
$D_b^F$. For more details,
we refer the reader to the supplementary material available from \cite{swh-dir-588b380}. 

\begin{table}[t]
\caption{The impact of rectifying a random forest by classification rules, in terms of F-score, G-mean, AUC score, size, and computation time.\label{tab:evolution2}}\vspace{0.3cm}

\centering
\resizebox{\linewidth}{!}{
\begin{tabular}{lrrrrrrrrrrrr}
\toprule
Dataset&$\mathcal{IF}$&$\mathcal{FF}$&$\mathcal{IG}$&$\mathcal{FG}$&$\mathcal{IA}$&$\mathcal{FA}$&$\mathcal{IN}$&$\mathcal{FN}$&$\mathcal{ID}$&$\mathcal{FD}$&$\mathcal{TR}$&$\mathcal{NR}$\\

\\

\midrule
arrowhead\_0\_vs\_1 & 0.88 & 0.90 & 0.85 & 0.87 & 0.85 & 0.87 & 980.2 & 44693.6 & 6.7 & 12.4 & 1.22e-01 & 2.2 \\
arrowhead\_0\_vs\_2 & 0.89 & 0.91 & 0.85 & 0.88  & 0.86  & 0.89 & 658.4 & 19252.8 & 5.3 & 13.4 & 2.78e-01 & 10.3 \\
arrowhead\_1\_vs\_2 & 0.78 & 0.87 & 0.79 & 0.85 & 0.80 & 0.86 & 1037.4 & 25144.4 & 6.7 & 14 & 3.40e-01 & 7.0 \\
australian & 0.73 & 0.74 & 0.76 & 0.78 & 0.77 & 0.78 & 11104.2 & 18725.2 & 16.3 & 18.4 & 5.30e+01 & 62.2 \\

balance & 0.93 & 0.95 & 0.93 & 0.95 & 0.93 & 0.95 & 8857 & 15349.6 & 12.1 & 14 & 4.76e+00  & 21.4 \\
biodegradation & 0.77 & 0.78 & 0.84 & 0.85  & 0.84 & 0.85 & 12611.8 & 398474.8 & 16.4 & 27.3 & 5.17e+02 & 90.0 \\
breast-tumor &  0.43 & 0.47 & 0.51 & 0.54 & 0.54 & 0.57 & 9513.8 & 24956.4 & 19.2 & 20.9 & 1.00e+01 & 15.4 \\
cleveland & 0.70 & 0.75 & 0.72 & 0.75 & 0.72 & 0.76 & 2518 & 11923.2 & 10.6 & 15.3 & 4.52e-01 & 19.5 \\
cnae & 0.86 & 0.94 & 0.88 & 0.96 & 0.88 & 0.96 & 2100 & 39332.2 & 23.7 & 28.8 & 4.46e-01 & 12.8 \\
compas &  0.57 & 0.58 & 0.62 & 0.63 & 0.63 & 0.64 & 77957 & 81260.4 & 19.7 & 20.9 & 2.86e+01 & 69.7 \\
contraceptive & 0.58  & 0.60  & 0.64  & 0.65  & 0.65 & 0.66  & 18479.8 & 39696.6 & 21.5 & 23.5 & 8.49e+00  & 53.0 \\
divorce & 0.95 & 0.96 & 0.95 & 0.96 & 0.95 & 0.96 & 442.6 & 7875.8 & 3.7 & 14.9 & 2.73e+00 & 19.3 \\
german & 0.98  & 0.98  & 0.02  & 0.02  & 0.49  & 0.50  & 5457 & 56676 & 15.1 & 24.9 & 5.15e+01  & 100.0 \\


\bottomrule
\end{tabular}
}
\end{table}

\begin{table}[t]
\caption{The impact of rectifying a decision tree by classification rules, in terms of F-score, G-mean, AUC score, size, and computation time.\label{tab:evolution_DT}}\vspace{0.3cm}

\centering
\resizebox{\linewidth}{!}{
\begin{tabular}{lrrrrrrrrrrrr}
\toprule
Dataset&$\mathcal{IF}$&$\mathcal{FF}$&$\mathcal{IG}$&$\mathcal{FG}$&$\mathcal{IA}$&$\mathcal{FA}$&$\mathcal{IN}$&$\mathcal{FN}$&$\mathcal{ID}$&$\mathcal{FD}$&$\mathcal{TR}$&$\mathcal{NR}$\\

\\
\midrule
arrowhead\_0\_vs\_1 & 0.82 & 0.86 & 0.80 & 0.85 & 0.81 & 0.85 & 20.4 & 39 & 5.4 & 6.3 & 1.06e-03 & 5.6 \\
arrowhead\_0\_vs\_2 & 0.83 & 0.86 & 0.80 & 0.84 & 0.81 & 0.84 & 12.8 & 20.2 & 4.2 & 5 & 1.38e-03 & 4.6 \\
arrowhead\_1\_vs\_2 & 0.74 & 0.78 & 0.74 & 0.77 & 0.74 & 0.78 & 19.2 & 33.2 & 5.3 & 6 & 7.24e-04 & 4.8 \\
australian & 0.70 & 0.74 & 0.73 & 0.77 & 0.73 & 0.77 & 141.4 & 5325.2 & 12.5 & 24.2 & 2.06e-01 & 28.0 \\
balance & 0.87 & 0.90 & 0.87 & 0.89 & 0.87 & 0.89 & 97 & 144.4 & 10 & 10.7 & 8.41e-03 & 15.9 \\
biodegradation & 0.62 & 0.69 & 0.73 & 0.77 & 0.73 & 0.78 & 189 & 97609.2 & 14.3 & 33.7 & 5.91e+00 & 74.3 \\
breast-tumor & 0.37 & 0.44 & 0.46 & 0.51 & 0.50 & 0.54 & 85.2 & 854.4 & 11.1 & 15.3 & 3.15e-02 & 12.8 \\
cleveland & 0.70 & 0.74 & 0.72 & 0.75 & 0.73 & 0.76 & 58.6 & 596 & 9 & 13.3 & 1.46e-02 & 12.9 \\
cnae & 0.90 & 0.95 & 0.95 & 0.98 & 0.95 & 0.98 & 54.4 & 3530.6 & 20.3 & 33.4 & 2.05e-01 & 34.9 \\
compas & 0.55 & 0.57 & 0.61 & 0.62 & 0.62 & 0.63 & 772.2 & 810 & 15.2 & 16.1 & 1.25e-01 & 49.5 \\
contraceptive & 0.57 & 0.60 & 0.62 & 0.64 & 0.62 & 0.65 & 448.8 & 689 & 17.6 & 17.6 & 9.15e-02 & 40.2 \\
divorce & 0.92 & 0.92 & 0.93 & 0.93 & 0.93 & 0.93 & 4 & 4.4 & 1.5 & 1.7 & 2.01e-04 & 2.3 \\
german & 0.96 & 0.98 & 0.27 & 0.29 & 0.54 & 0.56 & 64 & 1197.4 & 10.4 & 19.4 & 1.05e-01 & 48.9 \\

\bottomrule
\end{tabular}
}
\end{table}

\subsection{Empirical results}

Tables~\ref{tab:evolution2} and~\ref{tab:evolution_DT} give, respectively for the random forest model and the decision tree model, the initial and final (i.e., after rectification) average F-scores, G-mean scores, and AUC scores ($\mathcal{IF}$, $\mathcal{FF}$)$(\mathcal{IG}$,$\mathcal{FG})$ $(\mathcal{IA}$, $\mathcal{FA})$, the initial and final average numbers of nodes ($\mathcal{IN}$, $\mathcal{FN}$), and the  initial and final average depths ($\mathcal{ID}$, $\mathcal{FD}$).  $\mathcal{TR}$ represents the average cumulative time (in seconds) required to perform all rectifications. $\mathcal{NR}$ denotes the percentage of rectifications that led to change $F$.
We observe that, for the majority of the datasets used in the experiments, rectifying both random forests and decision trees with the mined classification rules leads to an increase in predictive performance, ranging from 1\% to more than 10\%. For the three metrics used, namely F-score, G-mean and AUC, an improvement is typically achieved. This improvement is small most of the time, but it can be very significant in some cases (see e.g., the {\tt arrowhead\_1\_vs\_2} and the {\tt cnae} datasets when a random forest is used).

The average computation times required to achieved the full  rectification of $F$ by $A_c$ were reasonable (less than $517$ seconds for the random forests and less than $5$ seconds for the decision trees).

The numbers of nodes and the depths increased for both classifiers, which is not a surprise given the way the rectification algorithm works. This increase can be significant and it may question the readability by humans of the rectified trees for some datasets. However, assuming that a tree-based model is ``human comprehensible'' when it contains at most $50$ nodes, only $4$ models out of the $26$ models considered in our experiments were already ``small enough'' at start to be viewed as ``human comprehensible'' ($\mathcal{IN} \leq 50$ for $4$ decision trees, only). And all of them of them remained ``small enough'' after the rectification step ($\mathcal{FN} \leq 50$ for $3$ decision trees).
The limit of $50$ nodes was systematically exceeded by the random forests that have been learned. 
Furthermore, preserving human interpretability was not an objective of our approach: what was expected instead (and actually achieved) was to keep the \emph{computational intelligibility} of the model \cite{DBLP:conf/kr/AudemardBBKLM21}, i.e., the ability to support efficiently a number of XAI queries. Especially, our experiments have shown that the ability to compute in an efficient way abductive explanations from tree-based models has been preserved after the rectification step.\footnote{It must also be kept in mind that if decision trees with a limited depth are more ``human comprehensible'', they are also less robust. Indeed, changing a few characteristics (no more than the depth of the tree) in an input instance (whatever it is) is enough to change the prediction made for this instance. See Proposition 5 in \cite{DBLP:journals/dke/AudemardBBKLM22} for details.} 

\begin{table}[h]
\caption{The impact of taking association rules into account on the size of abductive explanations given a random forest.\label{tab:statistics}}\vspace{0.3cm}

\centering
\begin{tabular}{lcccccccccccc}
\toprule
Dataset              & \multicolumn{3}{c}{100000} & \multicolumn{3}{c}{10000} & \multicolumn{3}{c}{1000} & \multicolumn{3}{c}{100} \\ 
                     & $\mathit{Red}$   & $\mathit{Ins}$ &       & $\mathit{Red}$   & $\mathit{Ins}$ &       & $\mathit{Red}$   & $\mathit{Ins}$ &       & $\mathit{Red}$   & $\mathit{Ins}$ &       \\ 
\midrule
arrowhead\_0\_vs\_1  & 29.83 & 100.00 &       & 10.52 & 97.01 &       & 5.18  & 96.12 &       & 2.29  & 80.62  &       \\ 
arrowhead\_0\_vs\_2  & 72.84 & 100.00 &       & 25.62 & 100.00 &       & 14.23 & 73.98  &       & 5.31 & 55.56  &       \\ 
arrowhead\_1\_vs\_2  & 55.73 & 100.00 &       & 16.11 & 98.08 &      & 4.84  & 92.59 &       & 2.36  & 64.96  &       \\ 
australian           & 3.68  & 43.90  &       & 1.70  & 23.57   &      &0.66 & 10.62 &        &0.11 & 1.92    \\ 

balance  & 0.00  & 0.00  &       & 0.00  & 0.00  &       & 0.00  & 0.00  &       & 0.00  & 0.00  &       \\ 
biodegradation & 37.38  & 99.60  &       & 5.36  & 90.30  &       & 1.48 & 46.90  &       & 0.13 & 9.70  &       \\
breastTumor & 19.62 & 91.96 &       & 0.55  & 6.32  &       & 0.00  & 0.00  &       & 0.00  & 0.00  &       \\ 
cleveland   & 0.59  & 19.09 &       & 0.27 & 8.43   &       & 0.22  & 4.33  &       & 0.17  & 2.65  &       \\ 
cnae     & 96.73 & 100.00&       & 41.74  & 98.85  &       & 21.42 & 95.05 &       & 16.75 & 92.30  &       \\ 
compas      & 7.37   & 37.40 &       & 0.00  & 0.00   &       & 0.00  & 0.00  &       & 0.00  & 0.00  &       \\ 
contraceptive    & 1.59  & 22.50 &       & 0.15  & 0.80 &       & 0.00  & 0.00  &       & 0.00  & 0.00  &       \\ 
divorce        & 82.63 & 100.00 &       & 82.88 & 100.00 &       & 43.09 & 100.00&       & 9.53 & 58.94 &       \\ 
german       &  0.78 & 28.30 &       & 0.4  & 9.50  &       & 0.12  & 9.31 &       & 0.09  & 0.60  &       \\ 

\bottomrule
\end{tabular}
\end{table}

The time required by the data mining algorithm to compute $A_c$ 
depends on the dataset and varies between $0.15$ and $1656.46$ seconds for the random forests. For decision trees, it ranges from $0.01$ to $0.46$ seconds.

\smallskip

Tables \ref{tab:statistics} and \ref{tab:statistics_decision_tree} give statistics about the evolution of the size of the smallest abductive explanations that have been found for the tree-based models $F$ considered at start, depending on the number of association rules that have been generated (up to $100$, $1000$, $10000$, and $100000$) and added to the initial domain theory $Th$ to get the extended domain theory $Th_e$. Such statistics are reported both for the random forests (Table \ref{tab:statistics}) and for the decision trees (Table \ref{tab:statistics_decision_tree}). In these tables, $\mathit{Red}$ represents the reduction achieved, i.e., on average over $100$ instances $\vec x$, the ratio between the size of the smallest UP-majoritary reason found for $\vec x$ using $Th$ minus the size of the smallest UP-majoritary reason found for $\vec x$ using $Th_e$, divided by the size of the smallest UP-majoritary reason found for $\vec x$ using $Th$.
$\mathit{Ins}$ represents on average over $100$ instances, the number of instances $\vec x$ for which the size of the smallest abductive explanation found has decreased when switching from $Th$ to $Th_e$. 

The results obtained show that for the two tree-based models considered in the experiments (decision trees and random forests), the reduction of the size of abductive explanations obtained by considering association rules  heavily varies with the dataset and with the number of association rules that are considered. Indeed, $\mathit{Red}$ is null or very small for some configurations and quite high for others. Of course, it increases with the number of association rules that are generated (since the logical strength of the theory cannot decrease when a rule is added to it). Similarly, the number $\mathit{Ins}$ of instances concerned by such a reduction heavily varies with the dataset and with the number of association rules that have been generated. It is null or very small for some configurations but reaches 100\% for other configurations when random forests are considered. Again, $\mathit{Ins}$  cannot decrease when the theory used is strengthened.


The time required by the data mining algorithm to compute 
the association rules that are not classification rules (i.e., those 
rules from $A \setminus A_c$)  
also depends on the dataset; it varies between 0.33 and 3600 seconds for the random forests and between 0.1 and 39.54 seconds for the decision trees. 
When deriving abductive explanations while taking advantage of association rules as a domain theory, experiments have not shown any huge extra computational cost, i.e., UP-majoritary reasons can be derived efficiently in practice (with computation times that are usually close to those required by majoritary reasons). More in detail, the computation time required to derive one UP-majoritary reason was  on average less than 1s,
whatever the dataset used and even when the domain theory has been completed with 100 000 rules). This lack of extra cost can be explained by the efficiency with which unit propagation can be achieved in practice. 

\begin{table}[h]
\caption{The impact of taking association rules into account on the size of abductive explanations given a decision tree.\label{tab:statistics_decision_tree}}\vspace{0.3cm}
%
\centering
\begin{tabular}{lcccccccccccc}
\toprule
Dataset              & \multicolumn{3}{c}{100000} & \multicolumn{3}{c}{10000} & \multicolumn{3}{c}{1000} & \multicolumn{3}{c}{100} \\ 
                     & $\mathit{Red}$   & $\mathit{Ins}$ &       & $\mathit{Red}$   & $\mathit{Ins}$ &       & $\mathit{Red}$   & $\mathit{Ins}$ &       & $\mathit{Red}$   & $\mathit{Ins}$ &       \\ 
\midrule
arrowhead\_0\_vs\_1  & 17.62 & 43.18 &       & 17.62 & 43.18 &       & 17.62 & 43.18 &       & 17.06 & 41.67  &       \\ 
arrowhead\_0\_vs\_2  & 11.12 & 26.76 &       & 11.12 & 26.76 &       & 11.12 & 26.76  &       & 11.12 & 26.76  &       \\ 
arrowhead\_1\_vs\_2  & 13.26 & 28.69 &       & 13.26 & 28.69 &      & 13.26 & 28.69 &       & 13.26 & 28.69  &       \\ 
australian           & 21.15  & 72.91  &       & 5.96  & 26.49  &      &1.82 & 10.95&        &0.25 & 2.04    \\ 

balance  & 0.00  & 0.00  &       & 0.00  & 0.00  &       & 0.00  & 0.00  &       & 0.00  & 0.00  &       \\ 
biodegradation & 38.68  & 85.51  &       & 19.93  & 68.52  &       & 3.20 & 17.72  &       & 0.55 & 3.70  &       \\
breastTumor & 17.09 & 65.29 &       & 11.59  & 54.03  &       & 3.42  & 19.78  &       & 0.39  & 1.77  &       \\ 
cleveland   & 4.29  & 24.73 &       & 4.29  & 24.73   &       & 0.48  & 2.88  &       & 0.17  & 0.66  &       \\ 
cnae     & 71.06 & 95.43&       & 71.06 & 95.43  &       & 41.40 & 78.26 &       & 3.48 & 28.90  &       \\ 
compas      & 3.76  & 19.10 &       & 0.00  & 0.00   &       & 0.00  & 0.00  &       & 0.00  & 0.00  &       \\ 
contraceptive    & 4.90  & 27.70 &       & 0.3  & 1.30 &       & 0.00  & 0.00  &       & 0.00  & 0.00  &       \\ 
divorce        & 19.54 & 33.33 &       & 19.54 & 33.33 &       & 19.54 & 33.33&       & 19.54 & 33.33 &       \\ 
german       & 5.38  & 20.63 &       & 5.38  & 20.63  &       & 2.19  & 7.63 &       & 0.00  & 0.00  &       \\ 

\bottomrule
\end{tabular}
\end{table}

\begin{table}[h]
\caption{Average computation time (in seconds, over 100 instances) for extracting a sufficient reason for an instance given a decision tree}
\label{tab:dt_avg_time}
\centering

\begin{tabular}{ l c c }
\toprule
Dataset & $\mathcal{T}_{\mathit{bef}}$ & $\mathcal{T}_{\mathit{aft}}$ \\
\midrule
arrowhead\_0\_vs\_1 & 0.00037 ± 0.00028 & 0.00042 ± 0.00024 \\
arrowhead\_0\_vs\_2 & 0.00030 ± 5.21e-05 & 0.00054 ± 0.00032 \\
arrowhead\_1\_vs\_2 & 0.00040 ± 0.00024 & 0.00045 ± 0.00028 \\
australian         & 0.00056 ± 0.00015 & 0.00157 ± 0.00095 \\
balance   & 0.00089 ± 0.00101 & 0.00061 ± 0.00017 \\
biodegradation      & 0.00045 ± 0.00025 & 0.00401 ± 0.00515 \\
breast-tumor\_0     & 0.00060 ± 8.15e-05 & 0.00136 ± 0.00112 \\
cleveland           & 0.00053 ± 9.90e-05 & 0.00112 ± 0.00107 \\
cnae\_0             & 0.00037 ± 0.00012 & 0.00139 ± 0.00098 \\
compas              & 0.00051 ± 0.00027 & 0.00041 ± 0.00020 \\
contraceptive\_0    & 0.00079 ± 0.00012 & 0.00116 ± 0.00099 \\
divorce             & 0.00028 ± 0.00014 & 0.00031 ± 0.00016 \\
german              & 0.00043 ± 5.56e-05 & 0.00066 ± 0.00025 \\
\bottomrule
\end{tabular}
\end{table}

\begin{table}[h]
\caption{Average computation time (in seconds, over 100 instances) for extracting a majoritary reason for an instance given a random forest}
\label{tab:rf_avg_time}
\centering

\begin{tabular}{ l c c }
\toprule
Dataset & $\mathcal{T}_{\mathit{bef}}$ & $\mathcal{T}_{\mathit{aft}}$ \\
\midrule
arrowhead\_0\_vs\_1 & 0.00108 ± 9.61e-05 & 0.03399 ± 0.01488 \\
arrowhead\_0\_vs\_2 & 0.00105 ± 0.00051 & 0.00984 ± 0.00510 \\
arrowhead\_1\_vs\_2 & 0.01427 ± 0.00317 & 0.02419 ± 0.02181 \\
australian         & 0.06901 ± 0.02262 & 0.08253 ± 0.02014 \\
balance             & 0.00880 ± 0.00190 & 0.01038 ± 0.00276 \\
biodegradation      & 0.09015 ± 0.02015 & 1.05798 ± 1.01648 \\
breast-tumor\_0     & 0.01202 ± 0.00430 & 0.04102 ± 0.01813 \\
cleveland           & 0.02420 ± 0.00565 & 0.06865 ± 0.02992 \\
cnae\_0             & 0.00273 ± 0.00033 & 0.01250 ± 0.00358 \\
compas              & 0.00994 ± 0.00592 & 0.00803 ± 0.00317 \\
contraceptive\_0    & 0.01224 ± 0.00394 & 0.01418 ± 0.00612 \\
divorce             & 0.00537 ± 0.00106 & 0.05882 ± 0.04120 \\
german              & 0.02064 ± 0.00688 & 0.03661 ± 0.01331 \\
\bottomrule
\end{tabular}
\end{table}

Tables~\ref{tab:evolution2} and~\ref{tab:evolution_DT} have shown that after rectification, the number of nodes in the models $F$ can increase significantly. Thus, it was important to assess whether this increase may have a significant impact on the time needed to compute abductive explanations for instances given $F$.
From such a perspective, Tables~\ref{tab:dt_avg_time} and~\ref{tab:rf_avg_time} report for the datasets considered in our experiments  average computation times (in seconds, over 100 instances) to generate one \textit{sufficient reason} for an instance given a decision tree and one \textit{majoritary reason} for an instance given a random forest.   $\mathcal{T}_{\mathit{bef}}$ denotes the time needed to compute a sufficient reason before the rectification of the model $F$, and $\mathcal{T}_{\mathit{aft}}$ the average time needed to compute a sufficient reason after the rectification of the model $F$ by $A_c$.
The domain theory considered here is the initial one $Th$.
From these tables, we can observe that while the time required to compute explanations increases after rectification, it remains very reasonable (in the worst case, less than 1s) whatever the model $F$.

\section{Other Related Work}

Though the issue of rule learning in a classification perspective has been considered for decades, our work departs significantly from previous approaches by the fact that it is also guided by the explanation issue.
Instead, the goal pursued by most of the previous approaches was to generate rule-based classifiers using classification rule mining. 
Furthermore, most of the time, in such previous approaches, tree-based models were not involved in addition to rule mining. 

An exception concerns the CBA system \cite{DBLP:conf/kdd/LiuHM98} based on Apriori algorithm \cite{DBLP:conf/vldb/AgrawalS94}. In the CBA system, classification rules $R$ are generated by decreasing confidence, then decreasing support unless preset minimal values are reached (in the reported experiments, 50\% for confidence $\mathit{minconf}$, and 1\% for the support $\mathit{minsup}$). A default class is considered as well so as to get a complete classifier whatever the number of rules generated.
The accuracy of the classifier is evaluated. 
The set of rules is pruned to get a minimal number of rules that cover the training data and achieve satisfying accuracy. In particular, every rule that does not enhance the accuracy of the classifier is removed.
\cite{DBLP:journals/kbs/LiuJLY08} improves the previous CBA approach in two directions. First, by considering minimal values $\mathit{minsup}$ that depend on the targeted class (this is important to get accurate rules when dealing with imbalanced datasets). Second, by making a competition of several classifiers on different segments of the training data. Each rule $R$ that is generated is thus used to select a subset of training instances, those covered by the body of the rule. 
Then several models are learned from the resulting set of instances and evaluated: $R$ is replaced by the model that minimizes the number of classification errors (provided that this number is lower than the number of errors coming from $R$). As remarked in \cite{DBLP:journals/kbs/LiuJLY08}, considering decision trees as one of the competitors is valuable since it makes it possible to generate deep trees (i.e., with paths corresponding to rules with many conditions) which can be necessary to get an accurate classification but can hardly be achieved by the rules generated by the CBA algorithm for combinatorial reasons. Thus, decision trees are used in this approach to improve a rule-based classifier, while in some sense a converse path is followed in our work.

In contrast to such works, rule mining is leveraged in our approach to improve a tree-based classifier as to inference and explanation. Unlike the previous approaches focused solely on the inference issue, only rules with 100\% confidence and non-null support are looked for. In our approach, for the inference task, primacy is given to classification rules over the tree-based classifier. A rectification algorithm is used to update the tree-based classifier with each classification rule that is generated. Finally, not only classification rules are extracted but other association rules are generated as well and exploited to improve the explanation task by strengthening the domain theory associated with the tree-based model.

As far as we know, the use of mined rules for generating more general abductive explanations has hardly been considered previously. As a notable exception, let us mention \cite{DBLP:conf/aaai/YuISN023}, which shows that taking into account mined rules is useful to diminish the size of sufficient reasons for decision lists (boosted trees and binarized neural networks are also mentioned). In this approach, rule induction is achieved using a specific MaxSAT-based algorithm to learn decision sets \cite{DBLP:conf/aaai/Ignatiev0S021}. Unsurprisingly, the empirical results presented in \cite{DBLP:conf/aaai/YuISN023} cohere with the ones pointed out in the previous section. The main differences between \cite{DBLP:conf/aaai/YuISN023} and our own work are actually threefold. First of all, 
\cite{DBLP:conf/aaai/YuISN023} also presents the impact of mined rules on the size of (subset-minimal) contrastive explanations (in theory, it is known that they can be lengthened) while only abductive explanations are considered in our work. Furthermore, in \cite{DBLP:conf/aaai/YuISN023} the mined rules are not used for the inference purpose.
Finally, the ML models considered in the two papers differ (\cite{DBLP:conf/aaai/YuISN023} focuses on decision list, boosted trees, and binarized neural networks, while we have considered decision trees and random forests).


\section{Conclusion}

In this paper, we have shown how to combine association rules derived using data mining techniques with decision tree and random forest classifiers.
Classification rules are exploited to tentatively enhance the predictive performance of the classifier at hand using a rectification operation.
Other association rules are leveraged to produce more general abductive explanations. Computational guarantees about the tractable generation of those explanations have been provided. 
An important feature of our approach is that at each step the user \emph{can keep the control} of the association rules from $A$ he/she is ready to consider (he/she can simply select those rules in which he/she is confident enough and filters out the other ones). 
Experiments have been conducted showing that the proposed approach is practical enough. Computation times remain reasonable, even for large datasets (see e.g., the results for 
\texttt{biodegradation} and \texttt{cnae} in the tables)
The empirical results have shown that the objective of improving the predictive performance of the classifiers can be reached, even if they are often modest for the inference task. 
On the contrary, they have also shown
that the reduction of the size of abductive explanations can be very significant.

This work calls for further research.  A perspective is to compile $Th_e$ in order to make it tractable for clausal entailment \cite{DarwicheMarquis02}. This would be an alternative to unit propagation as used in the proposed approach.
Provided that the compiled forms remain small enough
(which cannot be guaranteed in the general case), this could be a way to benefit from the full power of logical  entailment, and as a consequence, to get even more general explanations.
Experiments will be run to determine whether this is actually the case in practice.



\bibliographystyle{splncs04}
\bibliography{mybibfile}

\begin{thebibliography}{10}
\providecommand{\url}[1]{\texttt{#1}}
\providecommand{\urlprefix}{URL }
\providecommand{\doi}[1]{https://doi.org/#1}

\bibitem{DBLP:conf/vldb/AgrawalS94}
Agrawal, R., Srikant, R.: Fast algorithms for mining association rules in large
  databases. In: Proc. of VLDB'94. pp. 487--499 (1994)

\bibitem{DBLP:journals/inffus/ArrietaRSBTBGGM20}
Arrieta, A.B., D{\'{\i}}az, N., Ser, J.D., Bennetot, A., Tabik, S., Barbado,
  A., Garc{\'{\i}}a, S., Gil{-}Lopez, S., Molina, D., Benjamins, R., Chatila,
  R., Herrera, F.: Explainable artificial intelligence {(XAI):} concepts,
  taxonomies, opportunities and challenges toward responsible {AI}. Inf. Fusion
   \textbf{58},  82--115 (2020)

\bibitem{DBLP:journals/dke/AudemardBBKLM22}
Audemard, G., Bellart, S., Bounia, L., Koriche, F., Lagniez, J.M., Marquis, P.:
  On the explanatory power of {B}oolean decision trees. Data Knowl. Eng.
  \textbf{142},  102088 (2022)

\bibitem{DBLP:conf/aaai/AudemardBBKLM22}
Audemard, G., Bellart, S., Bounia, L., Koriche, F., Lagniez, J.M., Marquis, P.:
  Trading complexity for sparsity in random forest explanations. In: Proc. of
  AAAI'22. pp. 5461--5469 (2022)

\bibitem{DBLP:conf/kr/AudemardBBKLM21}
Audemard, G., Bellart, S., Bounia, L., Koriche, F., Lagniez, J.M., Marquis, P.:
  On the computational intelligibility of {B}oolean classifiers. In: Proc. of
  KR'21. pp. 74--86 (2021)

\bibitem{Audemardetal23}
Audemard, G., Lagniez, J.M., Marquis, P., Szczepanski, N.: On contrastive
  explanations for tree-based classifiers. In: Proc. of ECAI'23 (2023),
  117--124

\bibitem{swh-dir-588b380}
Audemard, G., Coste-Marquis, S., Marquis, P., Sabiri, M., Szczepanski, N.:
  Leveraging association rules for better predictions and better explanations
  (Oct 2025),
  \url{https://archive.softwareheritage.org/swh:1:dir:30fd5c30f7d31a6a33c056a3b28c345b0556e7fe}

\bibitem{DBLP:conf/ijcai/AudemardLMS24a}
Audemard, G., Lagniez, J., Marquis, P., Szczepanski, N.: Deriving provably
  correct explanations for decision trees: The impact of domain theories. In:
  Proc. of IJCAI'24. pp. 3688--3696 (2024)

\bibitem{DBLP:conf/ijcai/Coste-MarquisM21}
Coste{-}Marquis, S., Marquis, P.: On belief change for multi-label classifier
  encodings. In: Proc. of IJCAI'21. pp. 1829--1836 (2021)

\bibitem{coste-marquis23}
Coste-Marquis, S., Marquis, P.: Rectifying binary classifiers. In: Proc. of
  ECAI'23. pp. 485--492 (2023)

\bibitem{Dalal1992a}
Dalal, M., Etherington, D.W.: A hierarchy of tractable satisfiability problems.
  Information Processing Letters  \textbf{44}(4),  173--180 (1992),
  \url{ftp://ftp.cirl.uoregon.edu/pub/users/ether/tract-hier.ps.gz}

\bibitem{DarwicheH20}
Darwiche, A., Hirth, A.: On the reasons behind decisions. In: Proc. of ECAI'20.
  pp. 712--720 (2020)

\bibitem{DarwicheMarquis02}
Darwiche, A., Marquis, P.: A knowledge compilation map. Journal of Artificial
  Intelligence Research  \textbf{17},  229--264 (2002)

\bibitem{DBLP:conf/aaai/GorjiR22}
Gorji, N., Rubin, S.: Sufficient reasons for classifier decisions in the
  presence of domain constraints. In: Proc. of AAAI'22. pp. 5660--5667 (2022)

\bibitem{DBLP:conf/iui/Gunning19}
Gunning, D.: {DARPA's explainable artificial intelligence {(XAI)} program}. In:
  Proc. of IUI'19 (2019)

\bibitem{DBLP:journals/corr/abs-2107-01654}
Huang, X., Izza, Y., Ignatiev, A., Cooper, M.C., Asher, N., Marques{-}Silva,
  J.: Efficient explanations for knowledge compilation languages. CoRR
  \textbf{abs/2107.01654} (2021), \url{https://arxiv.org/abs/2107.01654}

\bibitem{DBLP:journals/ijar/HuangM24}
Huang, X., Marques{-}Silva, J.: On the failings of {S}hapley values for
  explainability. Int. J. Approx. Reason.  \textbf{171},  109112 (2024)

\bibitem{DBLP:journals/corr/abs-2012-11067}
Ignatiev, A., Narodytska, N., Asher, N., Marques{-}Silva, J.: On relating
  'why?' and 'why not?' explanations. CoRR  \textbf{abs/2012.11067} (2020)

\bibitem{IgnatievNM19}
Ignatiev, A., Narodytska, N., Marques{-}Silva, J.: Abduction-based explanations
  for machine learning models. In: Proc. of AAAI'19. pp. 1511--1519 (2019)

\bibitem{DBLP:conf/aaai/Ignatiev0S021}
Ignatiev, A., Lam, E., Stuckey, P.J., Marques{-}Silva, J.: A scalable two stage
  approach to computing optimal decision sets. In: Proc. of AAAI'21. pp.
  3806--3814 (2021)

\bibitem{joao-ijcai21}
Izza, Y., Marques-Silva, J.: On explaining random forests with {SAT}. In: Proc.
  of IJCAI'21. pp. 2584--2591 (2021)

\bibitem{DBLP:journals/jair/IzzaIM22}
Izza, Y., Ignatiev, A., Marques{-}Silva, J.: On tackling explanation redundancy
  in decision trees. J. Artif. Intell. Res.  \textbf{75},  261--321 (2022)

\bibitem{DBLP:conf/kdd/LiuHM98}
Liu, B., Hsu, W., Ma, Y.: Integrating classification and association rule
  mining. In: Proc. of KDD'98. pp. 80--86 (1998)

\bibitem{DBLP:journals/kbs/LiuJLY08}
Liu, Y., Jiang, Y., Liu, X., Yang, S.: {CSMC:} {A} combination strategy for
  multi-class classification based on multiple association rules. Knowl. Based
  Syst.  \textbf{21}(8),  786--793 (2008)

\bibitem{DBLP:journals/cacm/MarquesSilvaH24}
Marques{-}Silva, J., Huang, X.: Explainability is \emph{Not} a game. Commun.
  {ACM}  \textbf{67}(7),  66--75 (2024)

\bibitem{Miller19}
Miller, T.: Explanation in artificial intelligence: Insights from the social
  sciences. Artificial Intelligence  \textbf{267},  1--38 (2019)

\bibitem{Molnar19}
Molnar, C.: Interpretable Machine Learning - A Guide for Making Black Box
  Models Explainable. Leanpub (2019)

\bibitem{Nautaetal23}
Nauta, M., Trienes, J., Pathak, S., Nguyen, E., Peters, M., Schmitt, Y.,
  Schl\"{o}tterer, J., van Keulen, M., Seifert, C.: From anecdotal evidence to
  quantitative evaluation methods: A systematic review on evaluating
  explainable {AI}. ACM Comput. Surv.  \textbf{55}(13s) (2023)

\bibitem{scikit-learn}
Pedregosa, F., Varoquaux, G., Gramfort, A., Michel, V., Thirion, B., Grisel,
  O., Blondel, M., Prettenhofer, P., Weiss, R., Dubourg, V., Vanderplas, J.,
  Passos, A., Cournapeau, D., Brucher, M., Perrot, M., Duchesnay, E.:
  Scikit-learn: Machine learning in {P}ython. Journal of Machine Learning
  Research  \textbf{12},  2825--2830 (2011)

\bibitem{ShihCD18}
Shih, A., Choi, A., Darwiche, A.: A symbolic approach to explaining {B}ayesian
  network classifiers. In: Proc. of IJCAI'18. pp. 5103--5111 (2018)

\bibitem{DBLP:conf/aaai/YuISN023}
Yu, J., Ignatiev, A., Stuckey, P.J., Narodytska, N., Marques{-}Silva, J.:
  Eliminating the impossible, whatever remains must be true: On extracting and
  applying background knowledge in the context of formal explanations. In:
  Proc. of AAAI'23. pp. 4123--4131 (2023)

\bibitem{DBLP:journals/jar/ZhangS00}
Zhang, H., Stickel, M.E.: Implementing the {D}avis-{P}utnam method. J. Autom.
  Reason.  \textbf{24}(1/2),  277--296 (2000)

\end{thebibliography}
\newpage
\section*{Appendix: Proofs}
\smallskip

%
%
\noindent {\bf Proof of Proposition \ref{prop:greedy}}
\begin{proof}
Our greedy algorithm is as follows. For the case when $F(\vec x) = 1$,  
start with $t = t_{\vec x}$, and iterate over the literals $\ell$ of $t$ by checking whether $t$ deprived of $\ell$ 
is a UP-implicant given $Th$ of at least $\lfloor \frac{m}{2} \rfloor +1$ decision trees of $F$. 
If so, remove $\ell$ from $t$ and proceed to the next literal.
Once all literals in $t_{\vec{x}}$ have been examined, the final term $t$ is by construction a UP-implicant given $Th$ of 
a majority of decision trees in $F$, such that removing any literal from it would lead to a term that is no longer a UP-implicant given $Th$ of this majority. 
So, $t$ is by construction a UP-majoritary reason. The case when $F(\vec x) = 0$ is similar, by simply replacing each $T_i$ by its negation 
(which can be obtained in linear time by replacing every $0$-leaf in $T_i$ by a $1$-leaf and vice-versa).
This greedy algorithm runs in time polynomial in the size of the input $t_{\vec x}$, $F$ and $Th$ 
since on each iteration, checking whether $t$ is a UP-implicant given $Th$ of $T_i$ (for each $i \in [m]$) can be done in 
time polynomial in the size of $t$, $T_i$, and $Th$. 
Indeed, in order to decide whether $t$ is a UP-implicant given $Th$ of a decision tree $T_i$ of $F$, it is enough to test
that for every clause $\delta \in \cnf(T_i)$, $\delta$ contains a literal derivable by unit propagation
from $t \wedge Th$. The fact that this set can be derived in time linear in the size of $t \wedge Th$ completes the proof. 
\end{proof}
%
\end{document}